\documentclass{article}

\usepackage[utf8]{inputenc}
\usepackage[T1]{fontenc}
\usepackage{hyperref}
\usepackage{url}
\usepackage{graphicx}
\usepackage{amsmath, amssymb}
\usepackage{booktabs}
\usepackage{float}
\usepackage{siunitx}
\usepackage{subcaption}
\usepackage{algorithm}
\usepackage{algpseudocode}
\usepackage{setspace}
\usepackage{arxiv}
\newcommand{\headeright}[1]{}
\title{PlantSAM: An Object Detection-Driven Segmentation Pipeline for Herbarium Specimens}

\author{
  Youcef Sklab\thanks{Corresponding author: \texttt{youcef.sklab@ird.fr}}~\textsuperscript{\dag},
  Florian Castanet~\textsuperscript{\dag}, 
  Hanane Ariouat~\textsuperscript{\dag},
  Souhila Arib~\textsuperscript{\ddag},\\
  Jean-Daniel Zucker~\textsuperscript{\dag,\S},
  Eric Chenin~\textsuperscript{\dag},
  Edi Prifti~\textsuperscript{\dag,\S} \\
  \textsuperscript{\dag}IRD, Sorbonne Université, UMMISCO, F-93143, Bondy, France \\
  \textsuperscript{\ddag}Laboratoire ETIS, UMR 8051 CNRS, CY Cergy Paris Université, France \\
  \textsuperscript{\S}Sorbonne Université, INSERM, NutriOmique, Hôpital Pitié-Salpêtrière, Paris, France
}
\date{}
\renewcommand{\headeright}{}
\begin{document}
\maketitle

\begin{abstract}
Deep learning-based classification of herbarium images is hampered by background heterogeneity, which introduces noise and artifacts that can potentially mislead models and reduce classification accuracy. Addressing these background-related challenges is critical to improving model performance. We introduce \textit{PlantSAM}, an automated segmentation pipeline that integrates YOLOv10 for plant region detection and the Segment Anything Model (SAM2) for segmentation. YOLOv10 generates bounding box prompts to guide SAM2, enhancing segmentation accuracy. Both models were fine-tuned on herbarium images and evaluated using Intersection over Union (IoU) and Dice coefficient metrics. PlantSAM achieved state-of-the-art segmentation performance, with an IoU of 0.94 and a Dice coefficient of 0.97. Incorporating segmented images into classification models led to consistent performance improvements across five tested botanical traits, with accuracy gains of up to 4.36\% and F1-score improvements of 4.15\%. Our findings highlight the importance of background removal in herbarium image analysis, as it significantly enhances classification accuracy by allowing models to focus more effectively on the foreground plant structures.
\end{abstract}




\section{Introduction}

Plants are a fundamental component of Earth’s biodiversity, shaping ecosystems, forming the basis of trophic networks, and playing a crucial role in regulating the balance between carbon dioxide and oxygen \cite{savingplant2019}. However, biodiversity currently faces unprecedented threats from climate change and human activities \cite{herbarium2018, Digitization2017}. Natural history collections, particularly herbarium specimens, hold hundreds of years of data documenting the evolution of biodiversity and the environment \cite{roleNatural1996}. These collections provide invaluable insights into plant distribution, plant morphology, and environmental changes over centuries \cite{ariouat2023, Sohaib2020, Triki2022, herbarium2018}.

Recent digitization efforts have significantly enhanced the accessibility of these herbarium collections \cite{Sweeney2018LargescaleDO}. Institutions such as the National Museum of Natural History (MNHN)(\url{https://www.mnhn.fr/en}) in Paris have spearheaded large-scale digitization projects, making millions of specimens available online through platforms like ReColNat (\url{https://www.recolnat.org/en/}) and the Global Biodiversity Information Facility (GBIF) (\url{https://www.gbif.org}). These digitized collections enable researchers to conduct high-throughput analyses of plant traits, such as leaf size, shape, and organ counts \cite{OrangeYolo3}. They also support advanced studies in crop quality assessment \cite{jiang2020convolutional} and disease classification \cite{VT2022Disease} or even soils evolution \cite{grasso2024soils}. However, to efficiently navigate and utilize these massive datasets, automatic methods are essential for extracting comprehensive metadata and descriptive characteristics (traits) related to plant morpho-anatomy or traits specific to each specimen -e.g. its conservation state, or the presence / absence of particular organs \cite{Dhaka2021}.

Deep Learning (DL) has emerged as a promising approach for analyzing herbarium specimens \cite{Dhaka2021, ariouat2023, Sahraoui2023, ariouat2025, sklab2024, sklab-cari-2024, sklab2025SIMNET}. Over the last two decades, DL has driven profound advancements in artificial intelligence, leading to the development of innovative architectures such as convolutional neural networks (CNNs) and Vision Transformers (ViTs) \cite{Lecun98, AlexNet12, VGG15, Inception15, ResNet16, ViT21}. These models have achieved remarkable performance in computer vision tasks, including image classification and object detection \cite{Lecun98, AlexNet12, ViT21}. In botanical research, DL models have been successfully applied to identify and annotate plant organs in herbarium images \cite{CVPlant2018, Sohaib2020, ariouat2025}.

\begin{figure}
  \centering
  \fbox{\includegraphics[width=.8\linewidth]{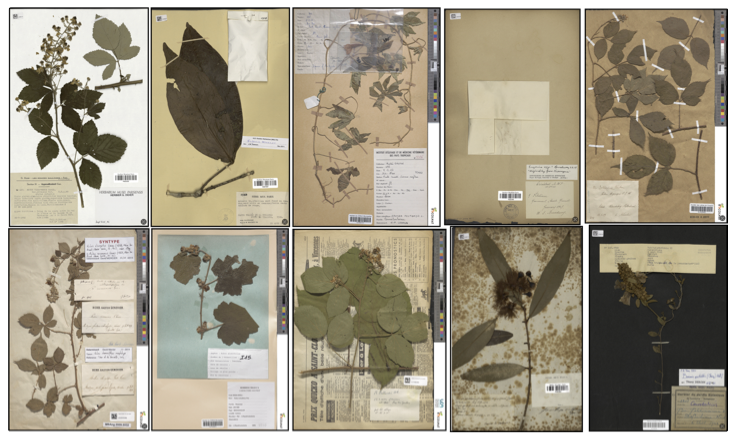}}
\caption{Examples illustrating the diversity in paper color, paper quality and the non-plant elements present in herbarium sheets (non-exhaustive) 
}
\label{fig:herbarium_example}
\end{figure}

Despite their potential, DL models often struggle with the heterogeneous backgrounds and artifacts present in herbarium images, such as labels, scale bars, overlapping plant structures, and aged paper textures (Figure \ref{fig:herbarium_example})\cite{sklab2025SIMNET, ariouat2025, sklab2024, sklab-cari-2024, CVPlant2018}. The presence of background noise can lead these models to learn irrelevant features, thereby reducing their performance in downstream tasks like trait classification and species identification \cite{Triki2021segmentation, Hussein2020, White2020, Fan2022, sklab2025SIMNET}. Therefore, isolating the foreground plant components (e.g., leaves, stems, flowers, and fruits) from non-plant background elements is essential to improve the accuracy and robustness of these models \cite{ariouat2025, sklab2024, sklab-cari-2024, sklab2025SIMNET}.


This task can be effectively addressed using image segmentation techniques \cite{ariouat2025,ariouat2023}. However, traditional segmentation models \cite{Triki2021segmentation, leafmachine2, Wilde2023, ariouat2023, Hussein2020, White2020, Triki2022DL, Fan2022} often struggle with these challenges, as they may focus on irrelevant features such as paper texture instead of plant morphology. Recent advancements in segmentation, particularly foundation models like the Segment Anything Model (SAM) \cite{kirillov2023segany} and its enhanced version SAM2 \cite{ravi2024sam2}, offer strong generalization capabilities across various segmentation tasks \cite{Yin2024SAMSO, Zhang2023PersonalizeSA, Zhao2023FastSA, Chen2024RobustSAMSA}. However, applying these models to herbarium images requires domain-specific fine-tuning and effective prompt generation, which can limit their scalability for large datasets \cite{ariouat2023, sklab2025SIMNET}.

In this work, we propose an automatic segmentation pipeline, called PlantSAM, that integrates the YOLOv10 object detection model \cite{wang2024yolov10} with SAM2 for segmenting plant regions. YOLOv10 generates bounding boxes as prompts, guiding SAM2 to segment plant structures. This approach removes the need for manual segmentation, enabling scalable segmentation across large herbarium datasets for downstream tasks. Our contributions are as follows:
\begin{itemize}
    \item Development of a novel segmentation pipeline combining YOLOv10 for prompt generation and SAM2 for plant segmentation.
    \item Fine-tuning of SAM2 and YOLOv10 on a curated dataset of herbarium images, addressing domain-specific challenges such as non-uniform backgrounds and complex plant structures.
    \item Evaluation of segmentation performance using Intersection over Union (IoU) and Dice coefficient, demonstrating significant improvements over UNet \cite{ariouat2025} and SAM2. 
    \item Assessment of the impact of segmentation on classification performance, showing accuracy improvements of up to 4.36\% and F1-score gains of 4.15\% across multiple botanical traits.
    \item Integration of the pipeline into a semi-automatic annotation tool to streamline mask refinement and improve the overall quality of the dataset.
\end{itemize}

\section{Related Work}

The analysis of plant specimens in herbarium images with deep learning methods has received significant attention, with a focus on segmentation techniques for tasks such as leaf counting, organ identification, and plant trait extraction \cite{Triki2021segmentation,leafmachine2,Wilde2023, Fan2022, ariouat2025, sklab2025SIMNET}.

Fan et al. \cite{Fan2022} explored segmentation and leaf counting in plants like \textit{Arabidopsis} and tobacco using UNet++ \cite{ZhouUnet++} with a ResNet50 backbone \cite{ResNet16}. While their approach effectively handled external elements such as soil and moss, it struggled with the complexities of herbarium images, which often feature intricate plant structures and highly variable backgrounds. Similarly, Triki et al. \cite{Triki2022} introduced a segmentation pipeline tailored for herbarium sheets, annotating elements like color palettes and labels. Their coarse segmentation method transformed these annotations into color-coded areas for an encoder-decoder model, assuming uniform transparent backgrounds. In contrast, our work addresses more complex herbarium specimens with diverse backgrounds. White et al. \cite{White2020} developed a segmentation model for ferns, employing Otsu’s thresholding to create training masks for a UNet-based architecture. However, their method was not designed to generalize to more diverse plant organs such as flowers, fruits, or seeds. Hussein et al. \cite{Hussein2020} applied DeepLabv3+ and Full-Resolution Residual Networks (FRRN-A) to herbarium images, achieving strong results in leaf and root segmentation. However, their work primarily focused on specific plant parts, whereas our approach generalizes to entire specimens, capturing all visible plant regions. Lee et al. \cite{lee2025multiple} introduced a novel neural network architecture specifically tailored for herbarium image segmentation, incorporating a multiple kernel-enhanced encoder to capture multiscale features. Their model employs convolutional blocks with diverse kernel sizes (1×1, 3×3, 5×5, and 7×7) to effectively learn both local and global patterns, addressing the structural complexity of herbarium specimens such as tangled roots and stems. Experimental results on a large dataset of various Viola species showed significant improvements in mIoU and precision compared to conventional models. While their approach advances segmentation accuracy through architectural design, our work complements this by focusing on trait-driven segmentation and multimodal integration for enhanced interpretability and generalization.

Thompson et al.~\cite{thompson2023identification} developed an object detection model based on YOLOv5 to identify non-plant components of herbarium sheets, such as institutional labels, handwritten annotations, or scale bars. Trained on annotated images from Melbourne University Herbarium (MELU) collection, their model showed promising generalization capabilities to other herbaria with minimal fine-tuning. Ott et al. \cite{ott2020ginjinn} introduced GinJinn, an object detection pipeline designed specifically to facilitate feature extraction from herbarium specimens using bounding-box–based models. GinJinn enables non-expert users to detect and extract botanical structures such as intact leaves with minimal programming effort. The pipeline was evaluated on images of various Leucanthemum species and demonstrated promising results in automatically identifying intact leaves from herbarium sheets. Ott and Lautenschlager \cite{ott2022ginjinn2} introduced GinJinn2, a general-purpose deep learning pipeline for object detection and segmentation in ecological and evolutionary studies. Their tool was applied to several biological datasets, including the segmentation of leaves in herbarium specimens. 

Sklab et al. \cite{sklab2025SIMNET} recently proposed Shape-Image Multimodal Network (SIM-Net), a novel architecture that enhances 2D image classification by integrating 3D point cloud representations inferred directly from RGB images. Designed for challenging datasets such as digitized herbarium specimens, SIM-Net combines a CNN-based encoder for visual features with a PointNet-based encoder for geometric cues derived from segmentation masks. By fusing both modalities into a shared latent space, SIM-Net improves robustness against noisy backgrounds and occlusions. Experimental results show substantial performance gains over ResNet101 and transformer-based models, demonstrating the value of integrating 3D structural reasoning into herbarium image classification.

Recent advances in foundation models like the Segment Anything Model (SAM2) \cite{kirillov2023segany,ravi2024sam2} have revolutionized image segmentation by leveraging pre-trained representations on large datasets. The Segment Anything Model (SAM) \cite{kirillov2023segany} is a foundation model for promptable image segmentation, introduced by Kirillov et al. (2023). It is designed to produce high-quality segmentation masks for a wide range of objects and scenes using various user-defined prompts. These prompts can be points, bounding boxes, or coarse masks, which guide the model to segment a specific object or region of interest. The key innovation lies in SAM’s ability to generalize to unseen objects without additional training, leveraging a powerful image encoder and prompt encoder coupled with a mask decoder.

SAM2 \cite{ravi2024sam2} builds on the original SAM by improving segmentation accuracy and efficiency, particularly for high-resolution or complex inputs. It introduces architectural refinements and enhanced prompt integration strategies, which make it more robust in cases with ambiguous boundaries, overlapping structures, or fine details —which are common challenges in herbarium images.

Both SAM and SAM2 are tailored to Promptable Visual Segmentation tasks, where the segmentation objective is conditioned on an input prompt. Unlike traditional semantic or instance segmentation models that operate in a fully automated fashion, promptable segmentation allows user or algorithmic control over which object(s) to segment, offering greater flexibility and scalability. Despite their versatility, the direct application of SAM and SAM2 to herbarium images has limitations. The lack of domain-specific fine-tuning and the reliance on manually generated prompts hinder scalability for large datasets. Moreover, SAM2 alone struggles with herbarium-specific challenges, including poorly defined contours, overlapping plant structures, and artifacts introduced by aging paper.  Ariouat et al. \cite{ariouat2025} demonstrated the potential of UNet with a ResNet101 backbone \cite{ResNet16} for herbarium segmentation, achieving IoU scores above 96\% on black and white background masks. However, this approach was less effective for specimens with thin stems or densely packed plant structures. 

\section{Plant Region Detection and Segmentation Pipeline}

\begin{figure}[H]
    \centering
    \fbox{\includegraphics[width=.7\linewidth]{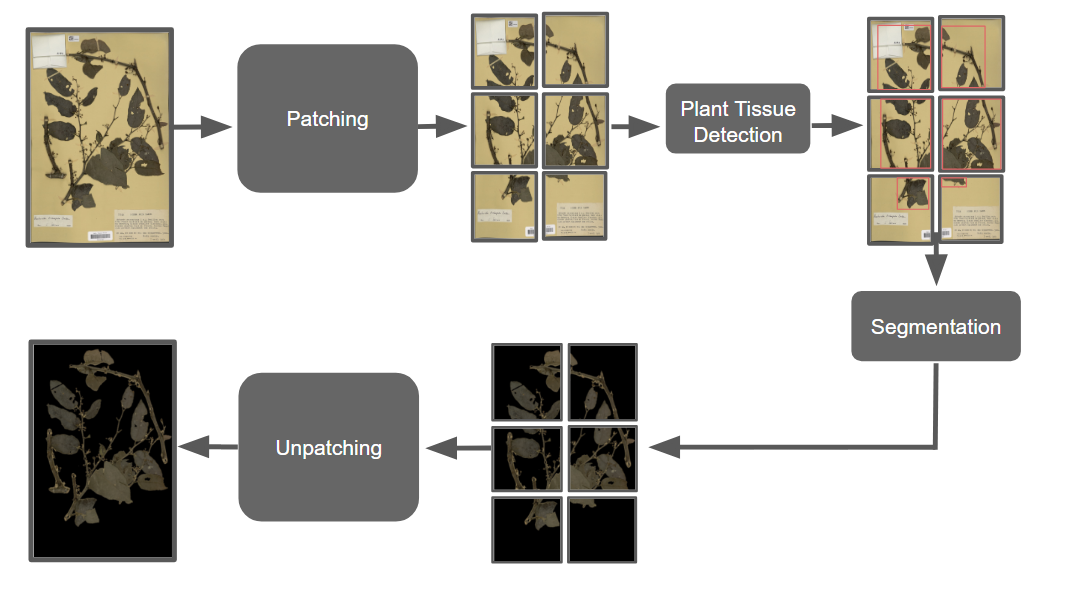}}
    \caption{Plant region detection and segmentation pipeline for herbarium images. Each image is divided into patches, processed by YOLOv10 for plant region  detection, and segmented by SAM. The segmented patches are then recombined to create a complete plant mask.}
    \label{fig:bbxopipeline}
\end{figure}

Our proposed pipeline (PlantSAM), as illustrated in Figure~\ref{fig:bbxopipeline}, comprises four phases: Patching, Plant Region Detection with YOLOv10, Segmentation with SAM, and Unpatching to recombine the segmented patches into a complete image mask. We evaluated both versions of SAM integrated into our pipeline: : the first version \cite{kirillov2023segany}, which we refer to here as SAM1, and SAM2. We used both SAM1 and SAM2 in our experiments to compare their performance with that of UNet \cite{ariouat2025} (with ResNet101 as backbone). Although SAM2 consistently outperformed SAM1 across all benchmarks, we retained SAM1 in our experiments for comparative purposes. We denote the pipeline using SAM1 as PlantSAM1, and the version using SAM2 as PlantSAM2. When referring to the method generically, we use PlantSAM, which encompasses both variants based on the integration of YOLOv10 with SAM.


\subsection{Image Preprocessing and Patching}

Each herbarium image, typically around 4000$\times$6000 pixels, is divided into smaller patches to preserve fine structural details and ensure efficient processing. This patching strategy mitigates resolution loss that occurs when handling the entire image and reduces computational load (Algorithm \ref{alg:image-processing}). YOLOv10 detects plant regions within each patch, generating bounding boxes that serve as prompts for SAM to segment the plant regions. The segmented patches are then recombined, with padding applied where necessary to maintain uniform dimensions, ultimately forming a complete image mask that isolates plant regions from the background.

\begin{algorithm}
\caption{Image processing}
\begin{algorithmic}[1]
    \State \textbf{Input:} Image path $path$
    \State \textbf{Output:} Image's patches
    \State $image \gets \text{load\_image}(path)$
    \State $image \gets \text{apply\_erosion}(image)$
    \State $image \gets \text{apply\_dilatation}(image)$
    
    \State $width \gets \text{get\_image\_width}(image)$
    
    \If {$\frac{width}{1024} > 3$}
        \State $patchs \gets \text{crop\_image}(image, 1024)$
    \ElsIf {$\frac{width}{512} > 3$}
        \State $patchs \gets \text{crop\_image}(image, 512)$
    \Else {}
        \State $patchs \gets \text{crop\_image}(image, 256)$
    \EndIf
    
    \State \textbf{Return} $patchs$
\end{algorithmic}
\label{alg:image-processing}
\end{algorithm}


Algorithm \ref{alg:image-processing} outlines the patch generation process. The image is first loaded and then preprocessed using morphological operations —specifically erosion followed by dilation (Lines 3–5). These operations aim to reduce noise from small artifacts or residual background elements that could hinder accurate segmentation. Erosion removes isolated bright pixels or thin protrusions, often corresponding to labels, paper artifacts, or dust, while dilation restores the main structures of the plant, ensuring they are not lost during erosion. Together, these operations improve the continuity and clarity of plant structures and ensure cleaner patches for downstream detection and segmentation. Next, the optimal patch size is selected based on the image width (1024, 512, or 256 pixels), allowing the algorithm to balance between context and resolution (Lines 6–14). Choosing an appropriate patch size is crucial: more patches risk including background noise, while fewer patches can lead to segmentation loss due to the fixed mask resolution of SAM (256×256). To further enhance segmentation quality, the patches are normalized to have zero mean and unit variance to match the input distribution expected by the pre-trained SAM model, ensuring consistent behavior during inference. Additionally, standard augmentations (random flips, rotations) are applied to improve generalization.

\subsection{Plant Region Detection with YOLOv10}

In our pipeline, we used YOLOv10 to detect plant regions within herbarium image patches, identifying components regions such as leaves, stems, flowers, and fruits. After the patching step, YOLOv10 processes each patch to produce bounding boxes around the plant structures.  The detection process handles various challenges, such as overlapping plant parts, background noise, and artifacts like labels, pins, or aged paper textures. YOLOv10's bounding boxes effectively capture the spatial extent of the plant regions while minimizing false positives from background elements. This capability is crucial for ensuring better segmentation in the subsequent step with SAM.


As illustrated in Figure~\ref{fig:bbxopipeline}, YOLOv10 scans each image patch and outputs bounding boxes that localize plant structures. The number and size of these boxes vary depending on the morphological complexity of the specimen. In patches containing multiple distinct plant areas, YOLOv10 generates several bounding boxes, enabling finer-grained segmentation. This multi-region detection strategy facilitates the isolation of individual plant components.


To train YOLOv10 for this task, we used the Plant Region Detection Dataset \cite{castanet2025plant}, which we constructed specifically for this work based on the segmented herbarium image dataset previously published in \cite{ariouat2025}. To build this dataset, we first split the segmented herbarium images into smaller patches, as described in Section 3.1, we then generated bounding boxes around the plant regions by detecting contiguous non-black pixel areas, as the background in the segmented images is uniformly black. This process enables automated and consistent annotation. The final dataset contains 19,078 image patches annotated with bounding boxes, split into 14,307 for training, 3,815 for validation, and 956 for testing. 

\subsection{Segmentation with SAM}

We explored two types of prompts for guiding the segmentation process: point prompts and bounding box prompts. Initial experiments with point prompts proved ineffective, primarily due to the absence of an automated method for accurately placing them. In particular, distinguishing between positive points (placed on the plant) and negative points (placed on the background) was non-trivial. Additionally, maintaining a proper balance between positive and negative points introduced further complexity. For these reasons, we chose to focus on bounding box prompts.


\begin{figure}[H]
    \centering
    \begin{minipage}[t]{0.47\linewidth}
        \centering
        \fbox{\includegraphics[width=.9\linewidth]{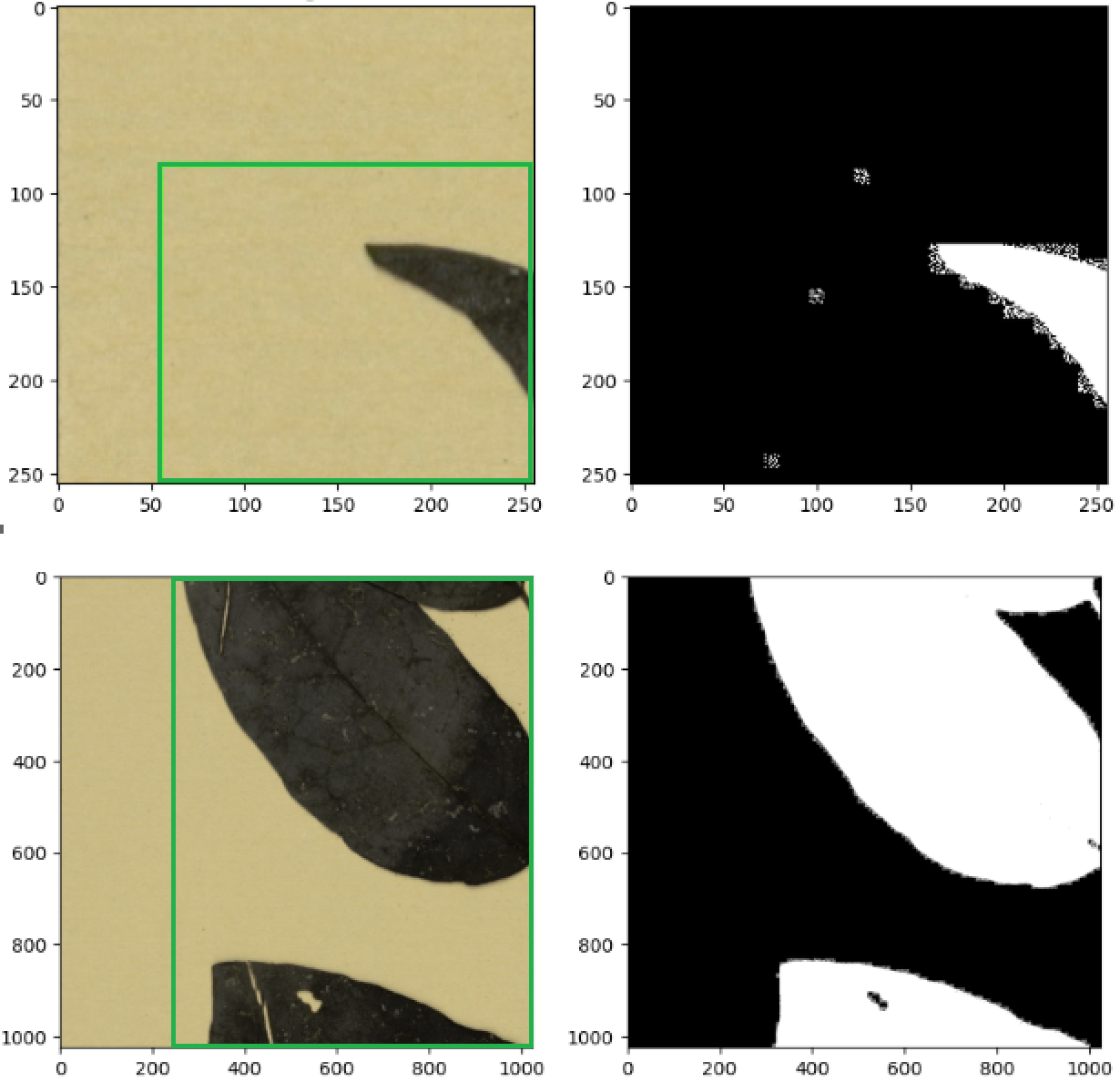}}
        \caption{Single-Box Strategy: A single bounding box encloses all plant regions in each patch, including residual background noise.}
        \label{fig:global-boxing}
    \end{minipage}
    \hfill
    \begin{minipage}[t]{0.47\linewidth}
        \centering
        \fbox{\includegraphics[width=.9\linewidth]{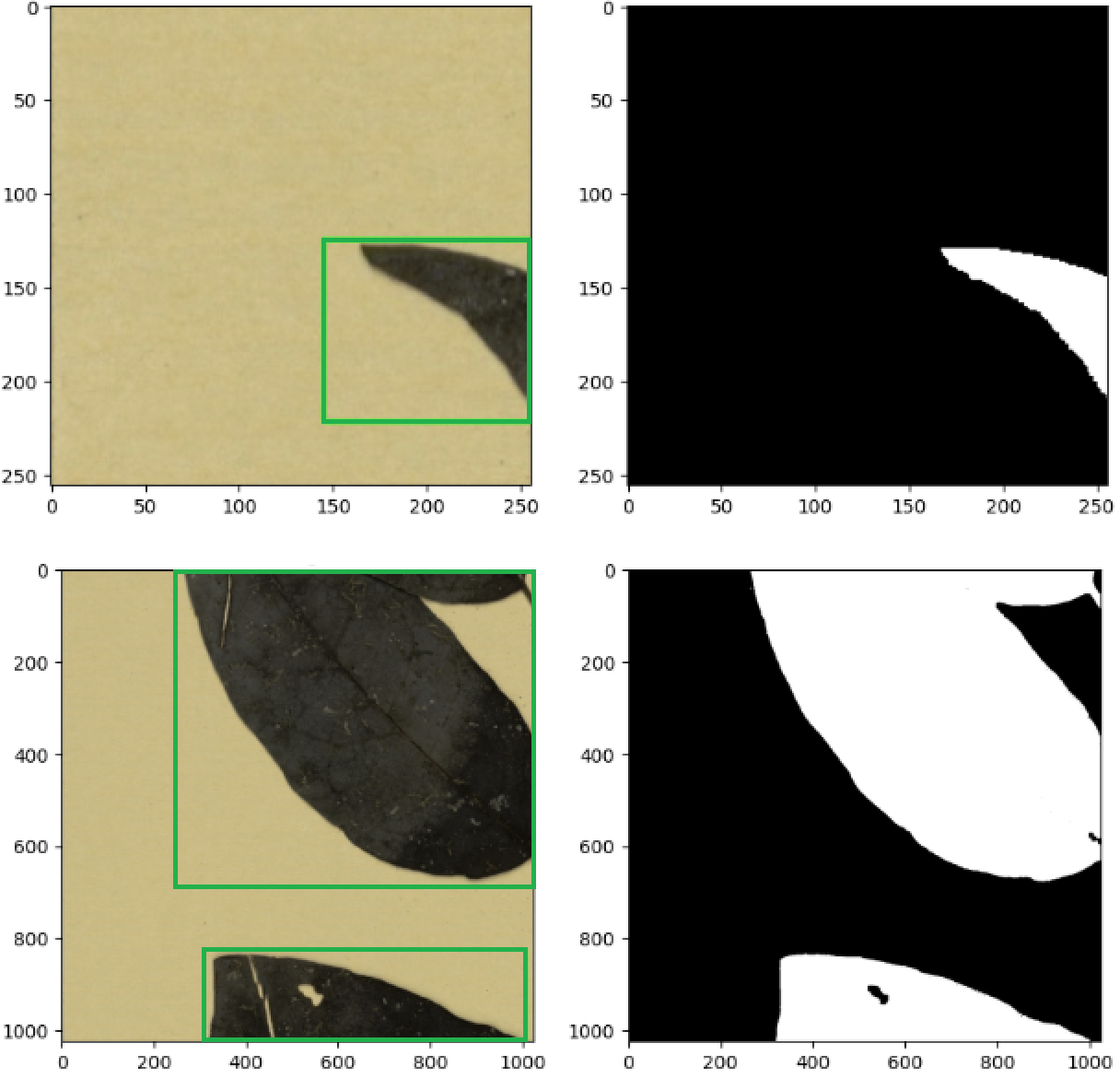}}
        \caption{Multi-Region Strategy: Multiple bounding boxes isolate distinct plant regions, reducing background noise and improving segmentation quality.}
        \label{fig:piece-boxing}
    \end{minipage}
\end{figure}

The bounding boxes, generated by YOLOv10, served as the input prompts for the SAM models. These prompts are essential for enabling SAM to perform better segmentation. As illustrated in Figure~\ref{fig:bbxopipeline}, once YOLOv10 identifies plant regions and outputs the corresponding bounding boxes, SAM generates segmentation masks for each patch. The segmented patches are then resized and reassembled to reconstruct the full-resolution segmentation of the original image. In our study, we investigated two bounding box prompting strategies:

\begin{itemize}
    \item \textbf{Single-Box Strategy}: A single bounding box is generated to enclose the entire plant structure within each patch, encompassing all visible plant regions. As illustrated in Figure~\ref{fig:global-boxing}, the box is created by identifying the outermost non-background pixels in the mask, forming a rectangle around the complete specimen. This strategy is computationally efficient and requires only one prompt per patch. However, it often captures surrounding background areas, which may introduce noise and reduce segmentation precision.

    \item \textbf{Multi-Region Strategy}: This approach generates multiple bounding boxes, each corresponding to a distinct connected component within the patch. As shown in Figure~\ref{fig:piece-boxing}, it isolates clusters of connected foreground pixels more precisely, allowing finer control over the segmentation. By focusing on localized regions, this strategy minimizes background inclusion and improves segmentation accuracy. However, it requires multiple prompts per patch, increasing computational cost and processing time.
\end{itemize}

\begin{figure}[]
    \centering
    \includegraphics[width=.55\linewidth]{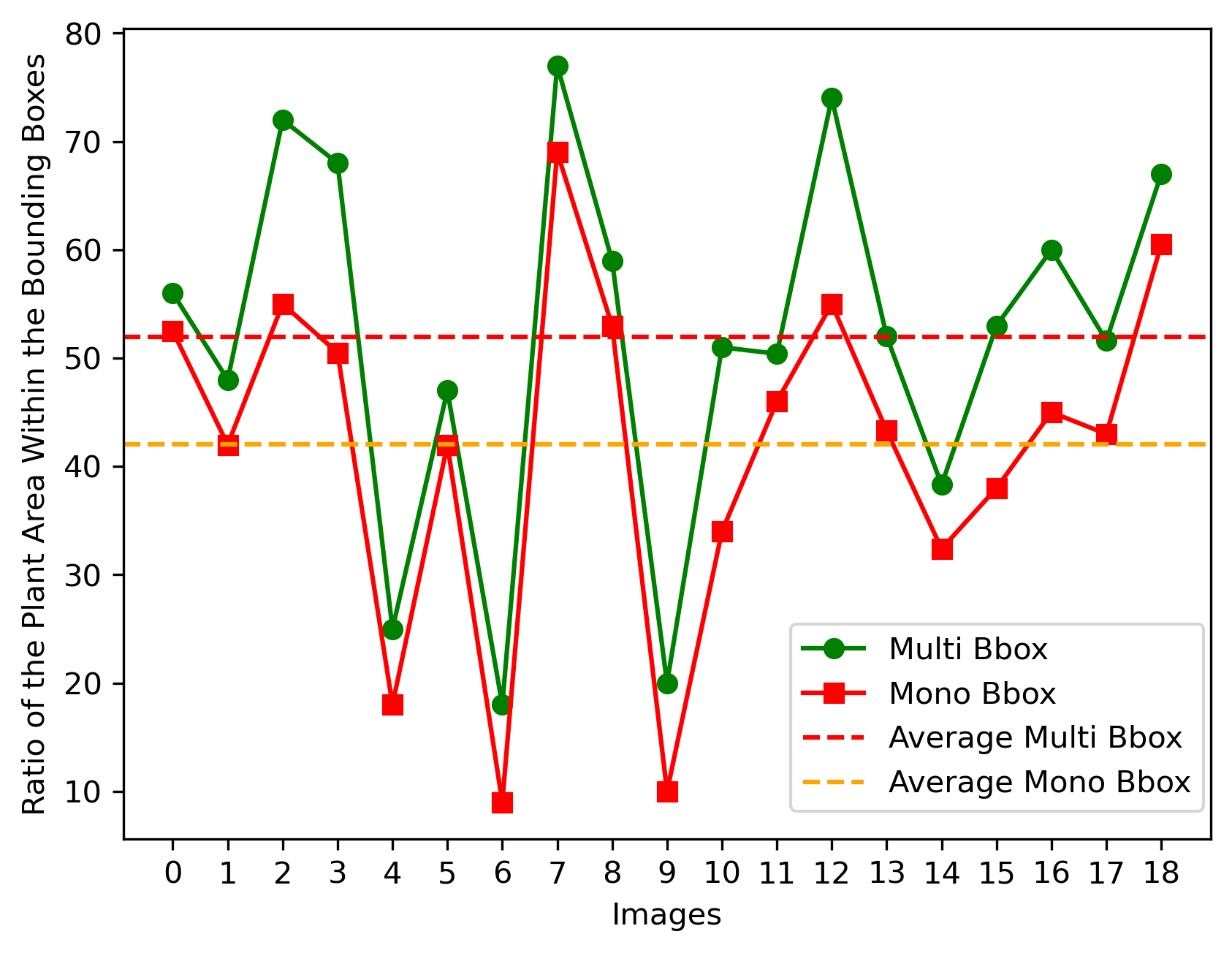}
    \caption{Comparison of segmentation between the multi-region strategy and the single-box strategy, based on the ratio of plant area to bounding box area. Each value represents the average ratio across all bounding boxes identified within a given image, using a total of 19 images. The multi-region strategy achieves a higher average ratio (51.96\%) compared to the single-box strategy (42.01\%), indicating improved segmentation performance by reducing background noise}
    \label{fig:yolocomparison}
\end{figure}

Morphological operations, such as erosion and dilation, were applied to enhance mask quality by removing problematic residual pixels and ensuring more accurate detection of plant structures  \cite{ariouat2025}. We conducted a comparative study on a set of 19 selected images for representative selection of our dataset. As shown in Figure~\ref{fig:yolocomparison}, the multi-region strategy reduces background inclusion within bounding boxes by approximately 10\% compared to the single-box strategy. Therefore, we incorporated it into our pipeline for better segmentation, particularly for complex or noisy herbarium images.


To fine-tune the SAM models, we used a curated subset \cite{castanet2025plant} of the \textit{Segmentation dataset} \cite{sklab2024herbarium} published in our previous work \cite{ariouat2025}. Specifically, we selected 1,476 herbarium images from the full dataset, chosen for their high segmentation quality based on visual inspection. The original dataset was constructed using a semi-automatic pipeline combining morphological operations and deep learning, as detailed in \cite{ariouat2025}.
We used specimens from 9 genera and 2 families for practical reasons regarding the images we had available.

\begin{figure*}
    \centering
    \begin{subfigure}[b]{0.19\linewidth}
        \centering
        \includegraphics[width=\linewidth]{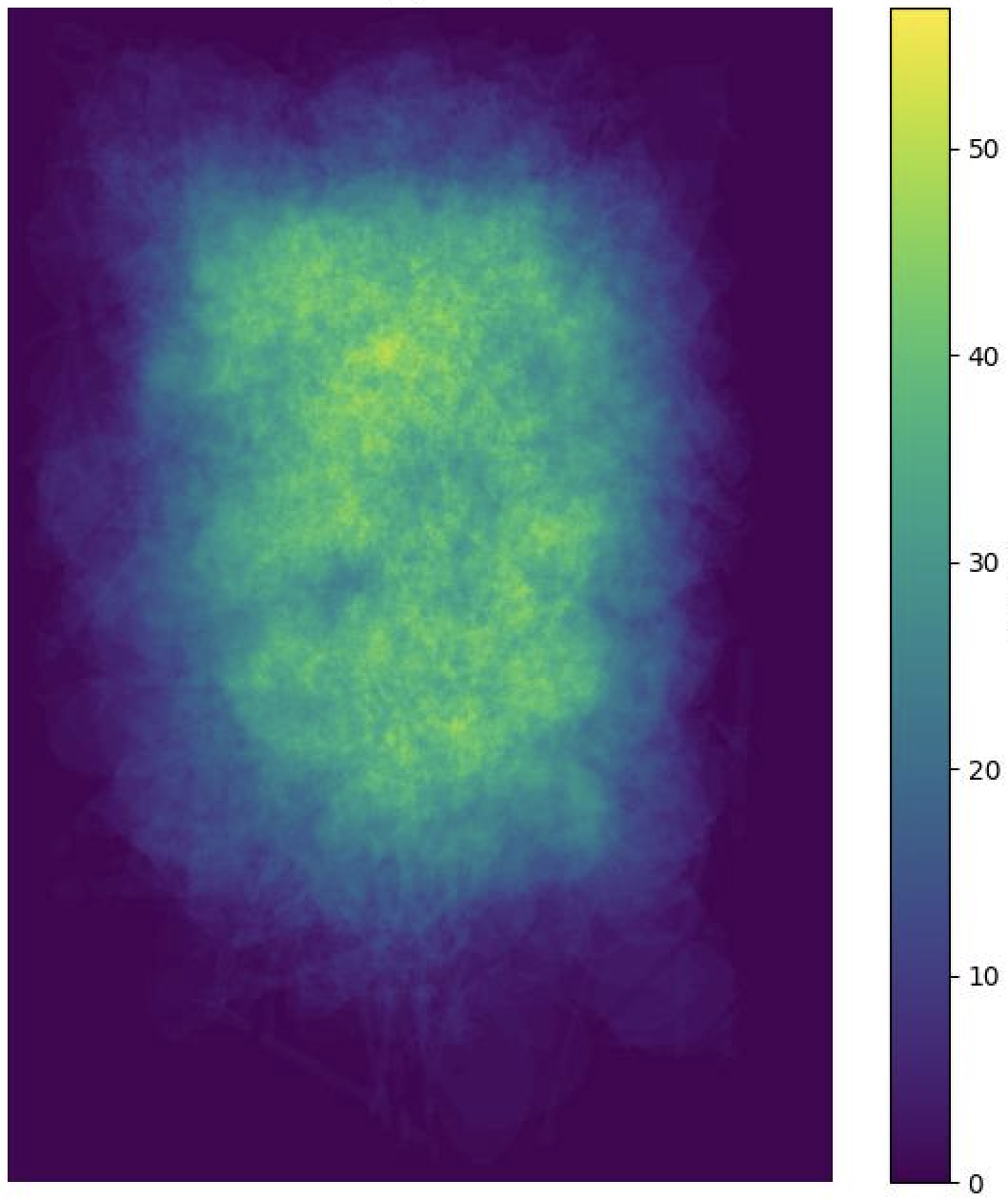}
        \caption{Rubus}
        \label{fig:rubus}
    \end{subfigure}
    \hfill
    \begin{subfigure}[b]{0.19\linewidth}
        \centering
        \includegraphics[width=\linewidth]{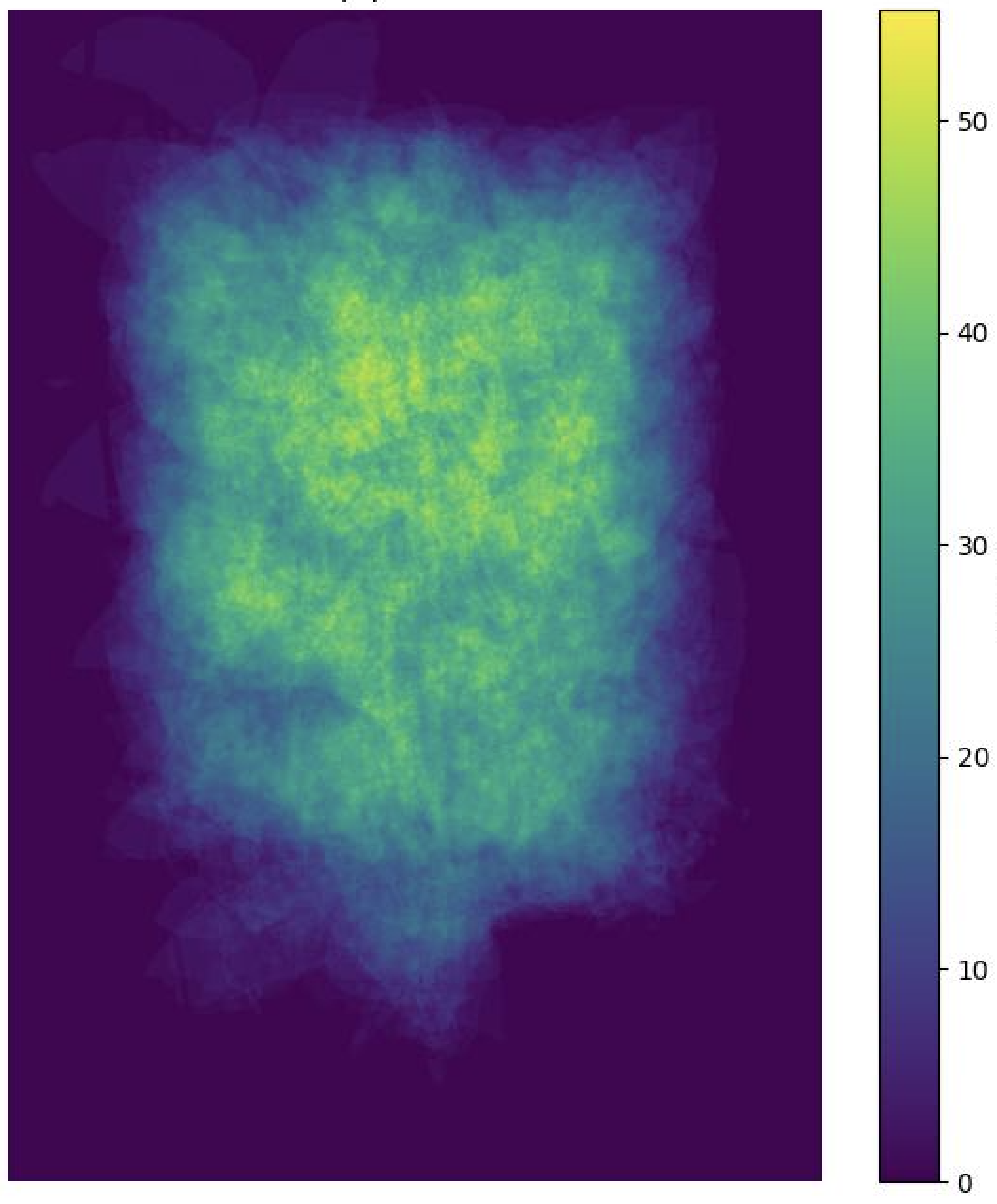}
        \caption{Amborella}
        \label{fig:amborella}
    \end{subfigure}
    \hfill
    \begin{subfigure}[b]{0.19\linewidth}
        \centering
        \includegraphics[width=\linewidth]{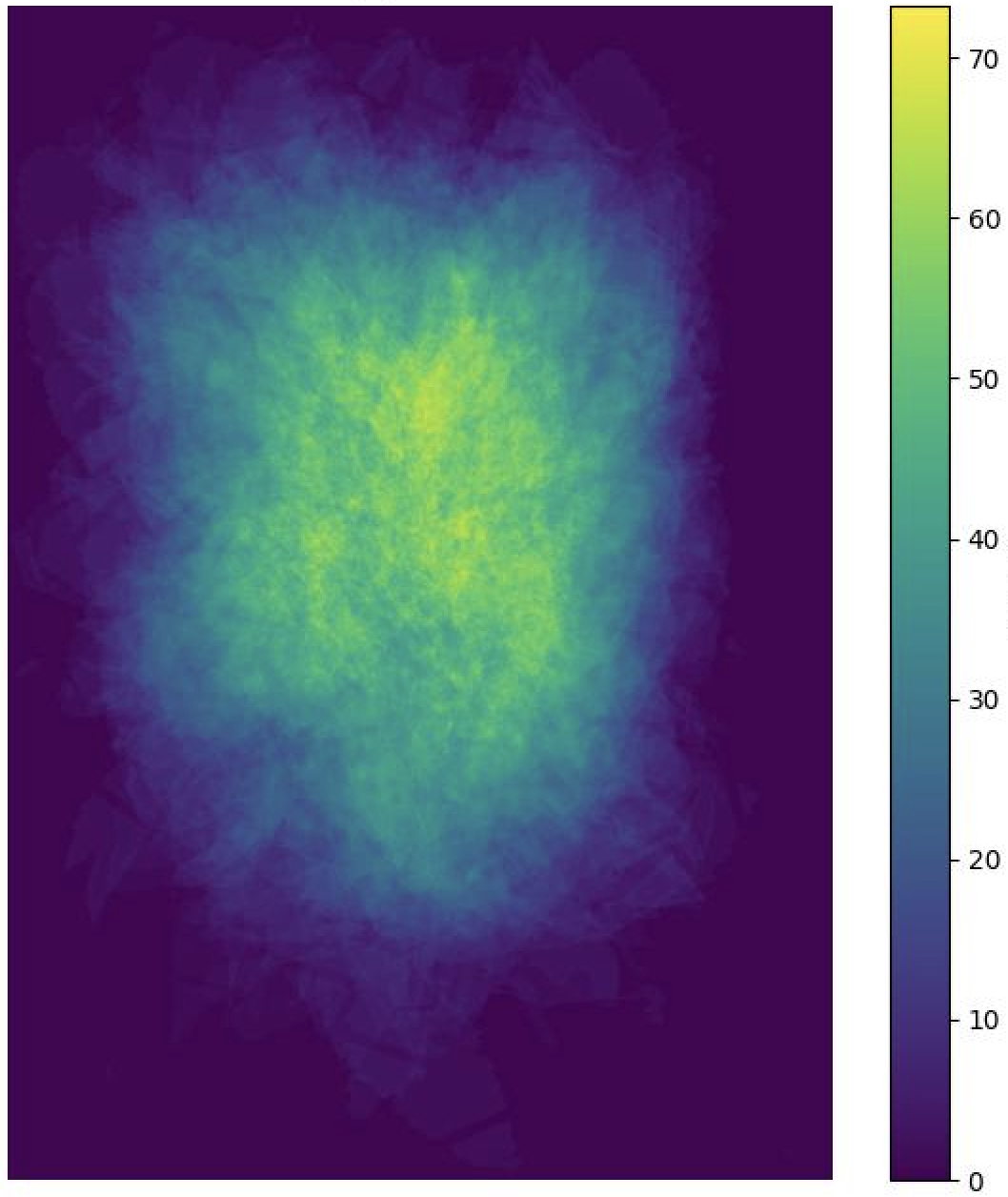}
        \caption{Castanea}
        \label{fig:castanea}
    \end{subfigure}
    \hfill
    \begin{subfigure}[b]{0.19\linewidth}
        \centering
        \includegraphics[width=\linewidth]{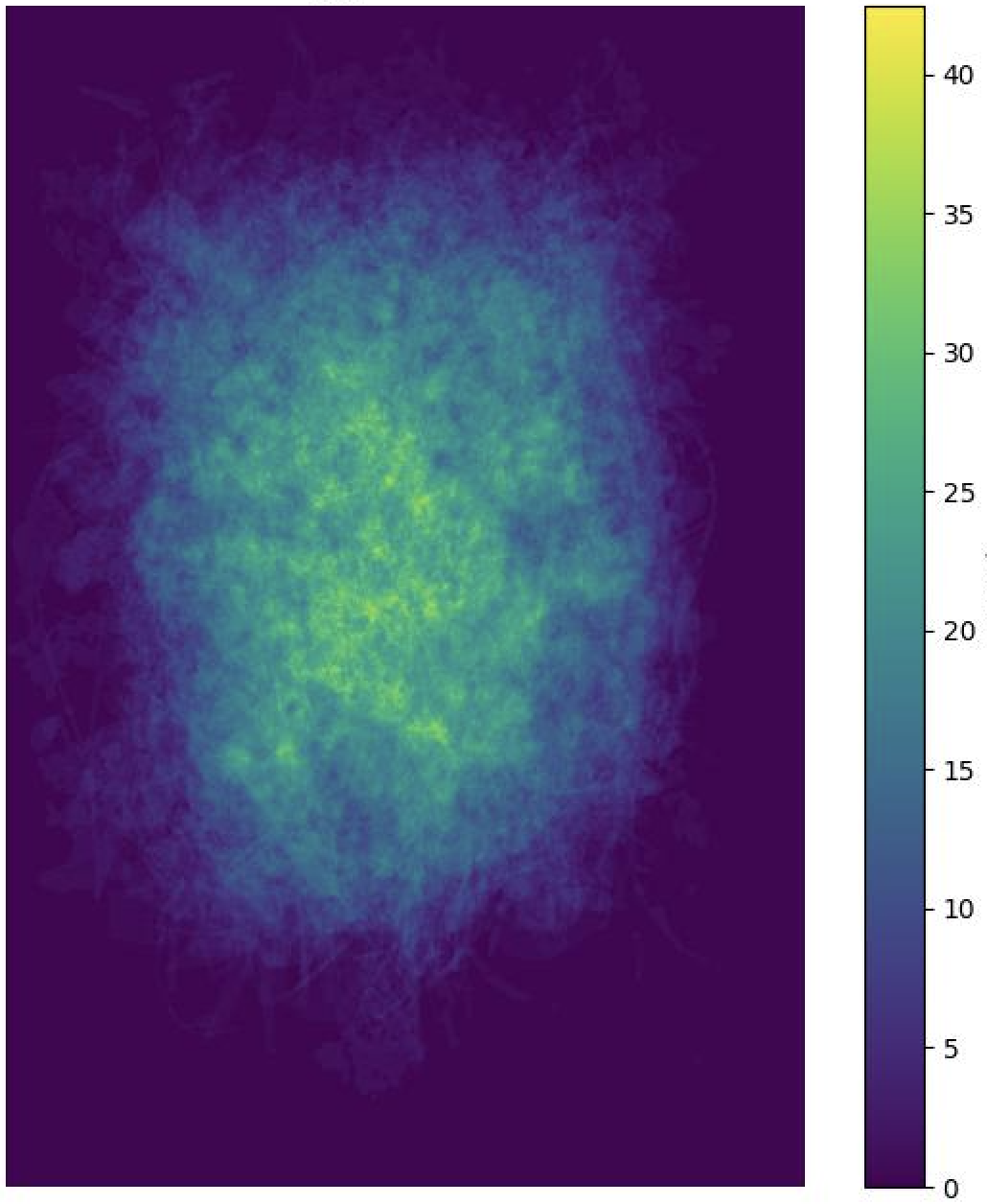}
        \caption{Desmodium}
        \label{fig:desmodium}
    \end{subfigure} 
    \hfill
    \begin{subfigure}[b]{0.19\linewidth}
        \centering
        \includegraphics[width=\linewidth]{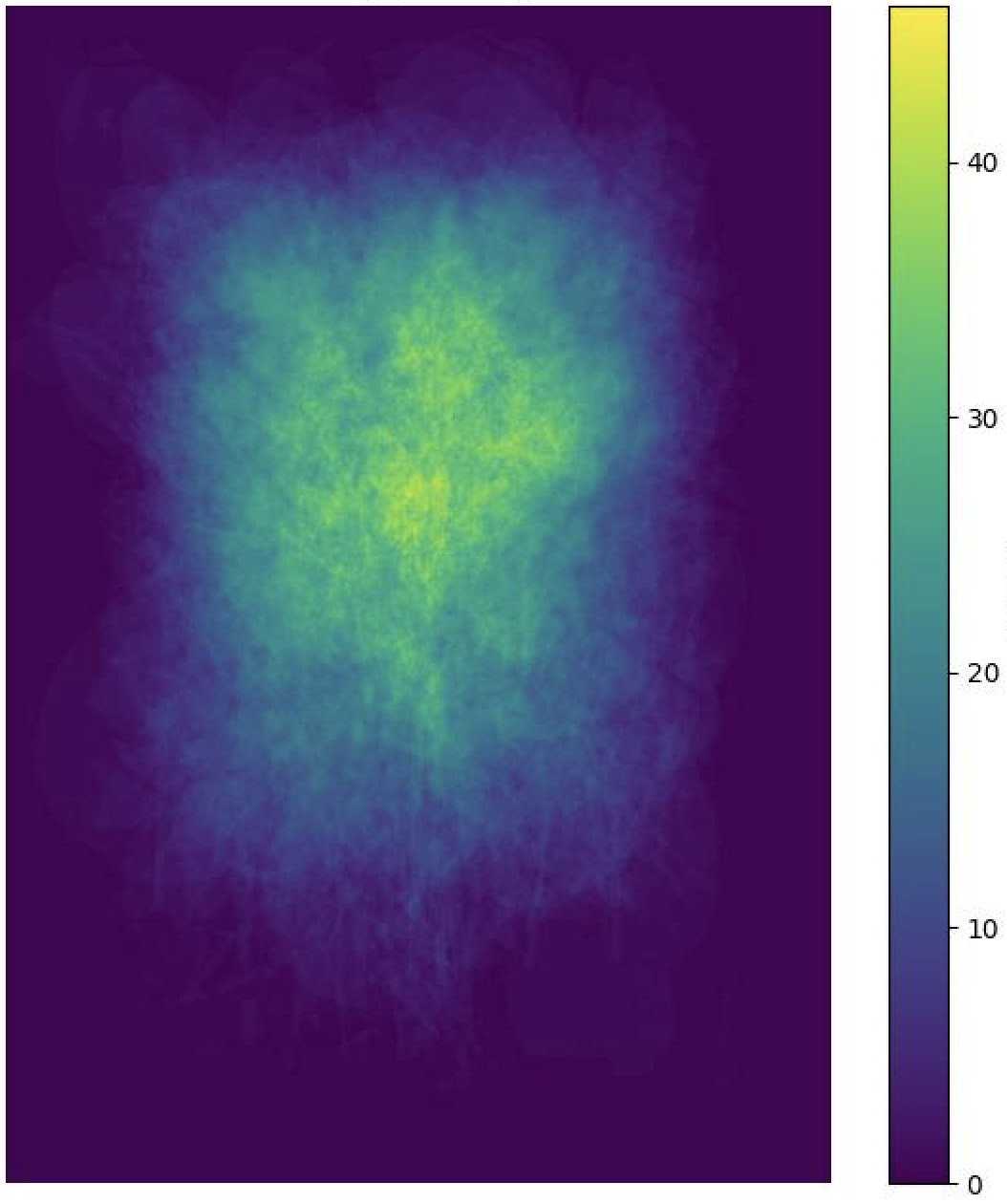}
        \caption{Eugenia}
        \label{fig:eugenia}
    \end{subfigure}

    \begin{subfigure}[b]{0.19\linewidth}
        \centering
        \includegraphics[width=\linewidth]{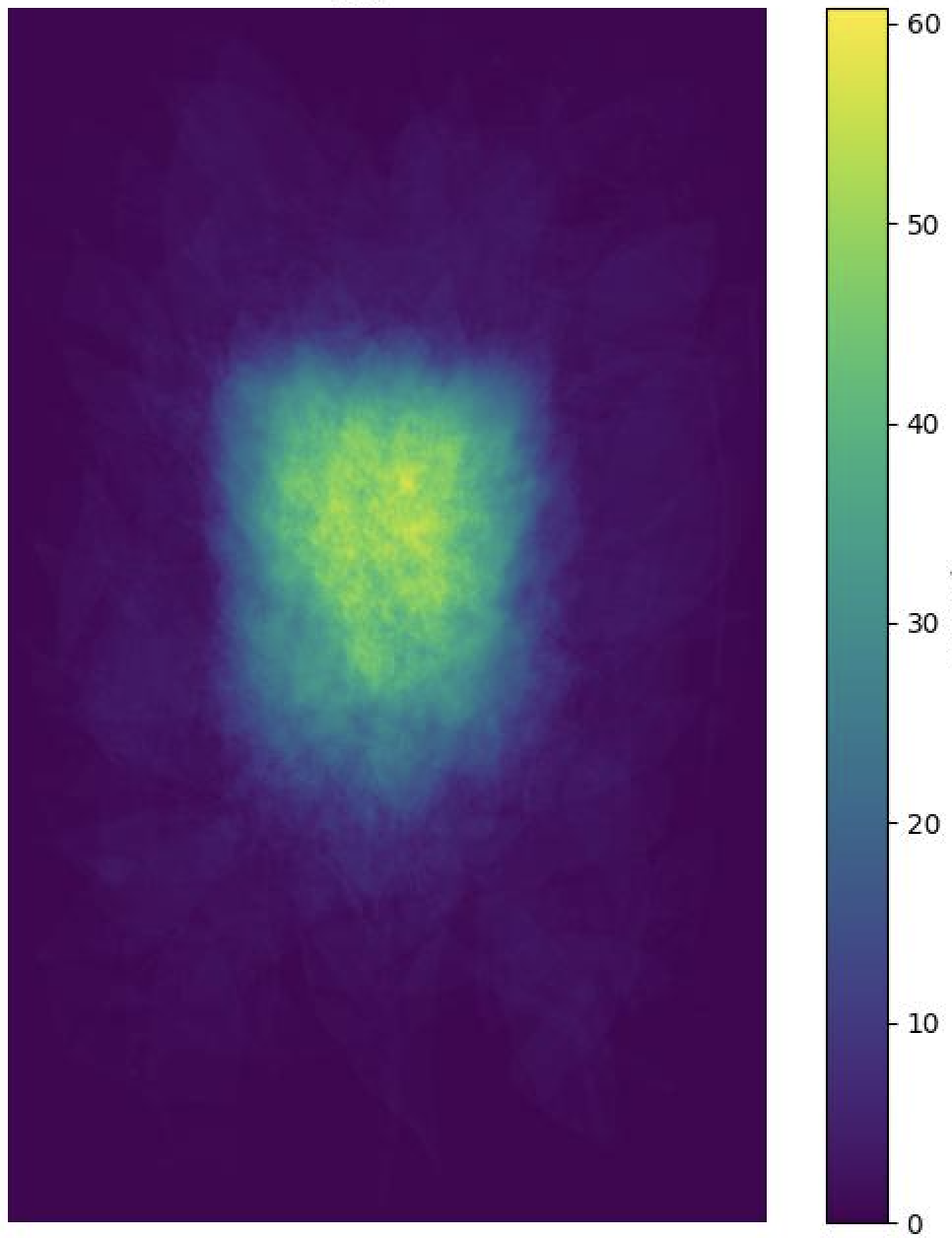}
        \caption{Laurus}
        \label{fig:laurus}
    \end{subfigure}
    \hfill
    \begin{subfigure}[b]{0.19\linewidth}
        \centering
        \includegraphics[width=\linewidth]{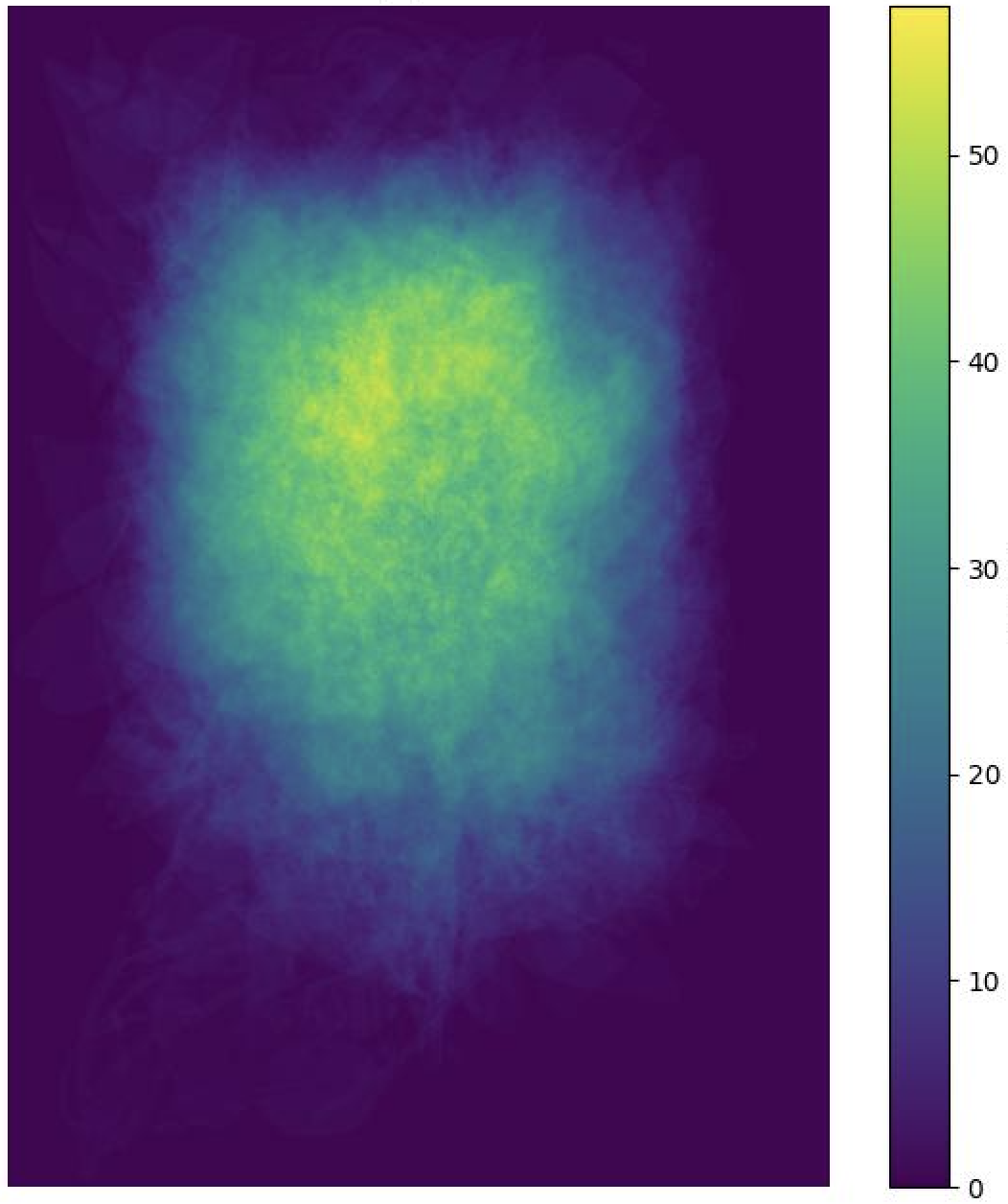}
        \caption{Litsea}
        \label{fig:litsea}
    \end{subfigure}
    \hfill
    \begin{subfigure}[b]{0.19\linewidth}
        \centering
        \includegraphics[width=\linewidth]{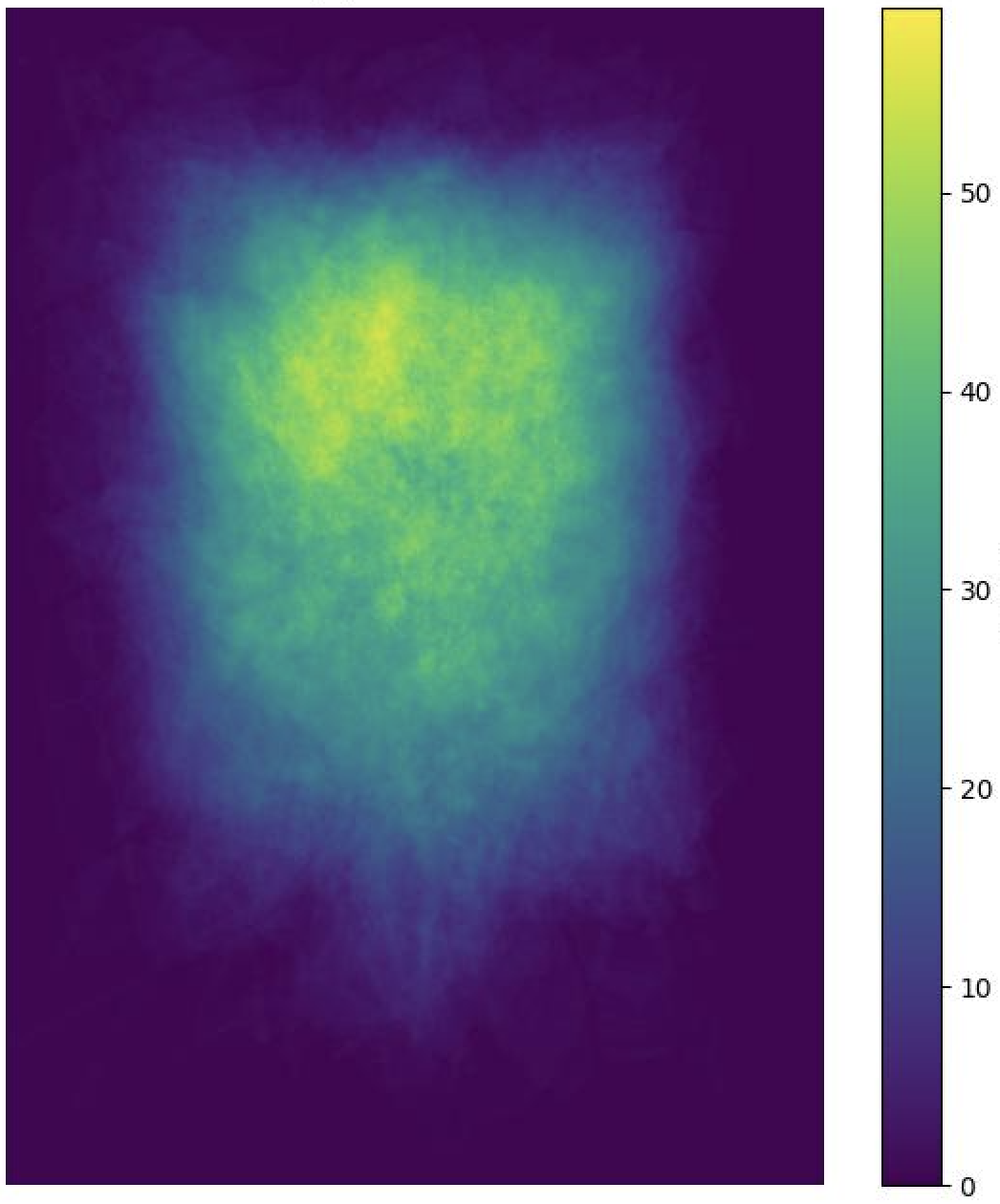}
        \caption{Monimiaceae}
        \label{fig:monimiaceae}
    \end{subfigure}
    \hfill
    \begin{subfigure}[b]{0.19\linewidth}
        \centering
        \includegraphics[width=\linewidth]{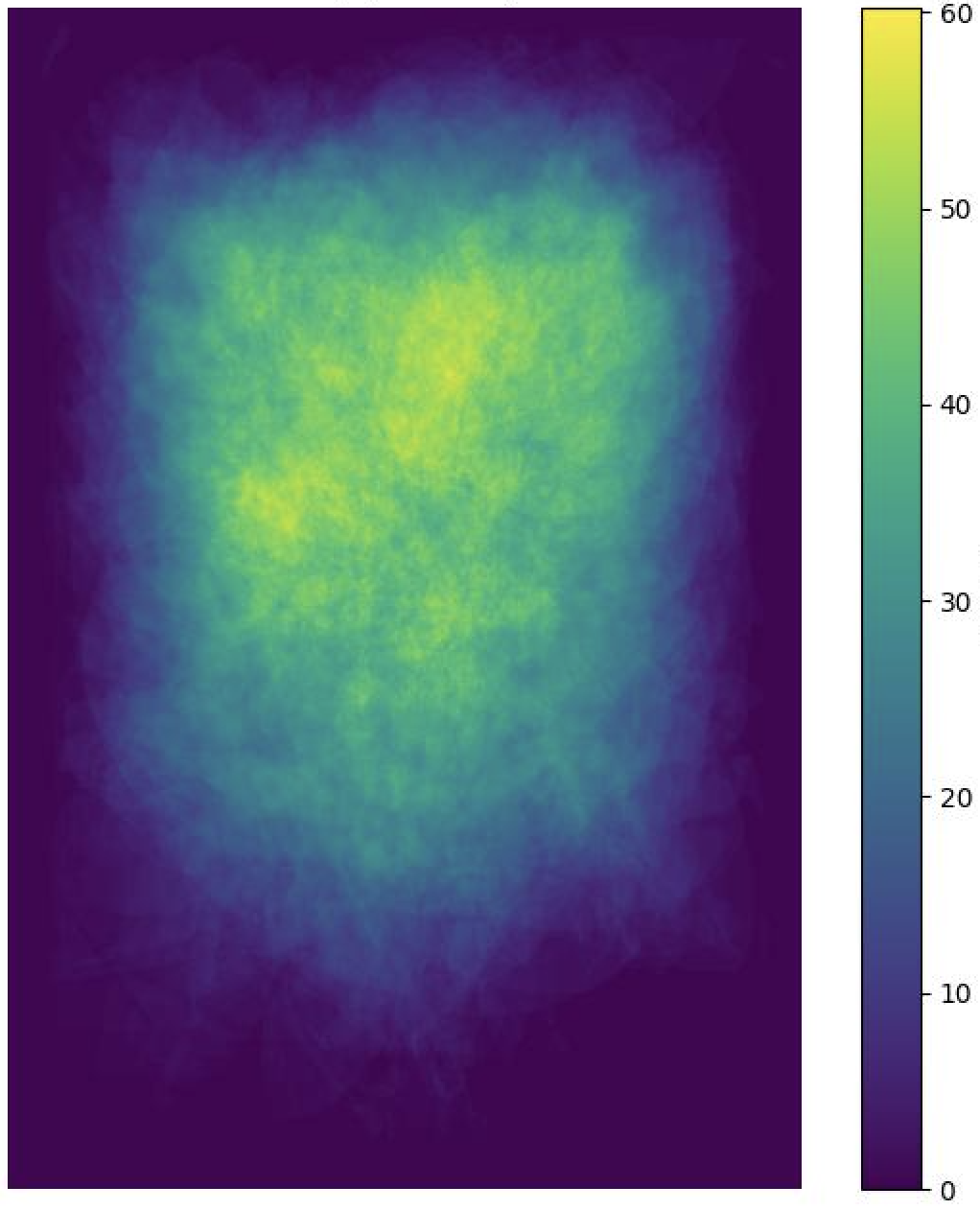}
        \caption{Magnolia}
        \label{fig:magnolia}
    \end{subfigure} 
    \hfill
    \begin{subfigure}[b]{0.19\linewidth}
        \centering
        \includegraphics[width=\linewidth]{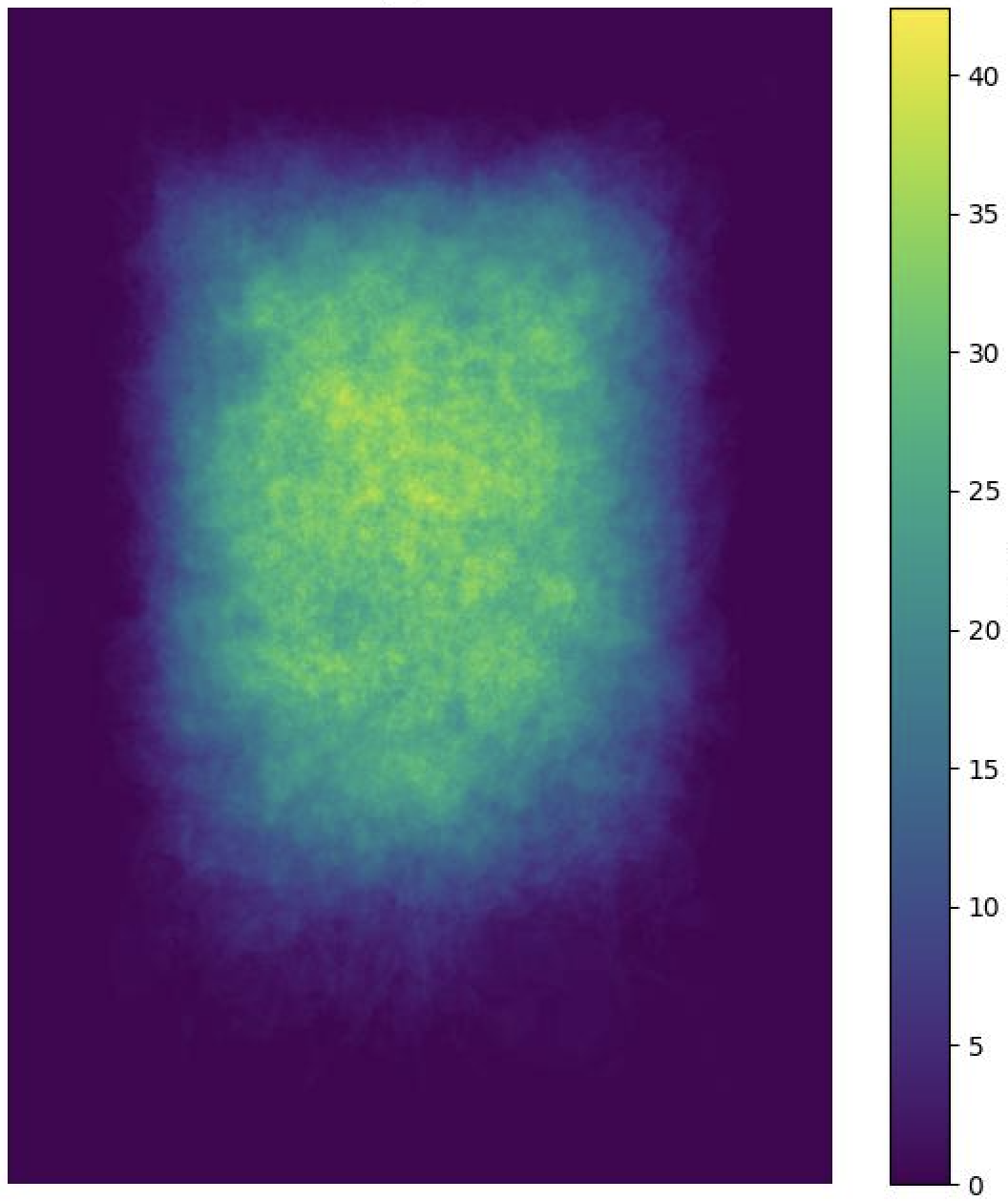}
        \caption{Ulmus}
        \label{fig:ulmus}
    \end{subfigure}

    \caption{Heat maps of the plant taxa, generated by combining and normalizing their masks. These visualizations highlight spatial distribution patterns, density, and surface coverage for each taxon.}
    \label{fig:heat-maps-plant-species}
\end{figure*}

We generated heat maps to analyze spatial distribution patterns, highlighting taxon-specific traits such as density and surface coverage (Figure~\ref{fig:heat-maps-plant-species}). Each visualization was centered around the largest specimen representative of the corresponding taxon. By aggregating and normalizing the segmentation masks across all images within a taxon, we produced color gradients that reveal the typical spatial positioning of plant structures on the herbarium sheet.


For instance, the heat map for \textit{Laurus} reveals a centralized and dense spatial distribution, suggesting that \textit{Laurus} specimens exhibit a compact structure, with plant elements predominantly clustered near the center of the herbarium sheet. In contrast, the heat map for \textit{Desmodium} shows a broader and more dispersed spatial distribution, indicating a less centralized organization. This variability aligns with the morphological traits of \textit{Desmodium}, known for its branching patterns and smaller, scattered leaves. The heat map for \textit{Magnolia} illustrates a highly centralized and dense distribution, reflecting the compact arrangement and large, broad leaves characteristic of \textit{Magnolia} specimens.

Table~\ref{tab:species_coverage} summarizes the percentage of surface area covered by different plant taxa in the segmentation dataset. We observe significant variation in coverage across taxa, reflecting differences in their morphological traits and specimen arrangements. For instance, Magnolia exhibits the highest average coverage (19.68\%), consistent with its large, broad leaves and compact structure. In contrast, Convolvulaceae has the lowest coverage (7.12\%), likely due to its more dispersed and slender growth habit. Other taxa, such as Castanea and Monimiaceae, show intermediate coverage levels, with values of 17.27\% and 14.80\%, respectively, illustrating the diversity in plant architecture present in the dataset. In some cases, over 90\% of the image consists of background rather than plant material. This finding quantitatively illustrates the challenge posed by high background content in herbarium images and highlights the need for segmentation as a preprocessing step. Removing background ensures that downstream models focus on relevant plant features.


\begin{table}[H]
\centering
\begin{tabular}{lrr}
\toprule
\textbf{Taxon}     & \textbf{Plant Coverage (\%)} & \textbf{Background Coverage (\%)} \\
\midrule
Magnolia       & 19.68    & 80.32   \\
Castanea       & 17.27    & 82.73   \\
Amborella      & 15.97    & 84.03   \\
Rubus          & 15.05    & 84.95   \\
Monimiaceae    & 14.80    & 85.20   \\
Litsea         & 14.17    & 85.83   \\
Eugenia        & 10.88    & 89.12   \\
Ulmus          & 10.87    & 89.13   \\
Desmodium      & 08.09    & 91.91   \\
Laurus         & 07.37     & 92.63   \\
Convolvulaceae & 07.12     & 92.88 \\
\hline
\end{tabular}
\caption{Average percentage of plant and background coverage for different plant taxa in the dataset. The genus Magnolia has the highest plant coverage (19.68\%), while the family Convolvulaceae has the lowest (7.12\%).}
\label{tab:species_coverage}
\end{table}



\subsection{Mask Reconstruction}

After processing all patches of an image, the segmented masks are reassembled to create a complete plant mask. As illustrated in Figure~\ref{fig:bbxopipeline}, this step, called "Unpatching", involves recombining the segmented patches to restore the original spatial context of the specimen. Reconstructed masks, adjusted for patching-related padding and refined using multi-region prompts, improve segmentation quality by minimizing background noise and preserving plant structures for downstream analysis. 


\section{Experiments}

In this section, we assess the performance of the segmentation pipeline under both ideal and non-ideal conditions, focusing on the comparison between UNet, SAM1, and SAM2. We evaluate the impact of combining YOLOv10 with SAM models using two datasets: a curated and annotated dataset representing ideal conditions, and an Out-Of-Distribution (OOD) dataset \cite{castanet2025digitised} featuring noisy and complex backgrounds.

\subsection{Experimental Setup}

We fine-tuned both SAM models (SAM1 and SAM2) on herbarium images using identical training conditions. UNet \cite{ariouat2025} was re-trained, while YOLOv10 was trained for plant region detection. 


To mitigate class imbalance during segmentation, we employed the Dice loss as the objective function for SAM1, SAM2, and UNet. All models were optimized using the Adam optimizer with an initial learning rate of \(10^{-5}\), reduced by a factor of \(10^{-1}\) every 10 epochs following a cosine annealing schedule. Training was conducted on the curated dataset of 1,476 herbarium segmentation masks \cite{castanet2025digitised}, with an 80/20 split between training and validation. We trained SAM1 and SAM2 for up to 80 epochs with a batch size of 1. For UNet \cite{ariouat2025}, training was limited to a maximum of 100 epochs, with early stopping based on validation Dice scores.

YOLOv10 was trained for 250 epochs on the plant region detection dataset \cite{castanet2025plant}, which was split into 14,307 training images, 3,815 for validation, and 956 for testing. All input images were resized to 640$\times$640 pixels. The best-performing YOLOv10 model achieved a mean Average Precision (mAP) of 0.959 at an IoU threshold of 0.5. All models were trained and evaluated on NVIDIA A100 GPUs.

\subsection{Segmentation Performance}

In this section, we evaluate the performance of our segmentation pipeline, focusing on PlantSAM1, PlantSAM2, and UNet. The evaluation consists of three key parts: quantitative analysis using standard metrics, generalization tests under challenging conditions, and the impact of segmentation on traits classification.

\subsubsection{Quantitative Evaluation}

We use two primary metrics for the quantitative evaluation: \textbf{Intersection over Union (IoU)} and \textbf{Dice coefficient} to compare PlantSAM1, PlantSAM2, and UNet across multiple plant taxa.

\begin{itemize}
    \item \textbf{Intersection over Union (IoU)} measures the overlap between the predicted segmentation mask and the ground truth mask. It is defined as the ratio of the intersection of these two masks to their union. A higher IoU score indicates a greater degree of overlap, reflecting more accurate segmentation.
    \begin{equation}
    \text{IoU} = \frac{\text{Intersection (Predicted Mask} \cap \text{Ground Truth Mask)}}{\text{Union (Predicted Mask} \cup \text{Ground Truth Mask)}}
    \end{equation}

    \item \textbf{Dice Coefficient} quantifies the overlap between the predicted and ground truth masks. It is computed as twice the intersection divided by the sum of the areas of both masks. The Dice coefficient ranges from 0 to 1, where a higher value indicates better segmentation performance.
    \begin{equation}
    \text{Dice Coefficient} = \frac{2 \times \text{Intersection}}{\text{Area of Predicted Mask} + \text{Area of Ground Truth Mask}}
    \end{equation}
\end{itemize}

The quantitative evaluation was conducted using a separate test dataset of 333 herbarium images \cite{castanet2025digitised}. The results are summarized in Tables~\ref{tab:comparaison} and~\ref{tab:comparaison-dice}. PlantSAM2 consistently outperformed both PlantSAM1 and UNet, achieving an average IoU of 0.94 and a Dice coefficient of 0.97. Notable improvements were observed for taxa such as \textit{Convolvulaceae} and \textit{Desmodium}, where PlantSAM2 demonstrated increases in IoU of 5.67\% and 6.81\%, respectively, compared to UNet. However, for taxa like \textit{Litsea}, gains were minimal, indicating variability in boundary-sensitive segmentation performance. It is also worth noting that PlantSAM1 was outperformed by UNet in nearly all the results presented in Table~\ref{tab:comparaison-dice} for the Dice coefficient metric.

\begin{table}[H]
\centering
    \begin{tabular}{lcccccc}
        \toprule
        \textbf{Taxon}  & \textbf{UNet} & \textbf{PlantSAM1}  & \textbf{Delta}  & \textbf{PlantSAM2}&\textbf{Delta}& \textbf{Number of Images}\\
        \midrule
        Amborella  & 0.9005 
        & 0.9387& +3.82 & 0.9432&+4.27 & 05 \\
        Castanea  & 0.9325 
        & 0.9394& +0.69 & 0.9425&+1.00 & 36 \\
        Convolvulaceae  & 0.8222 
        & 0.8698& +4.76 & 0.8789&+5.67 & 25 \\
        Desmodium  & 0.8337 
        & 0.9007& +6.70 & 0.9018&+6.81 & 11 \\
        Eugenia  & 0.9078 
        & 0.9275& +1.97 & 0.9324&+2.46 & 40 \\
        Laurus  & 0.9420 
        & 0.9558& +1.38 & 0.9569&+1.49 & 53 \\
        Litsea  & 0.9343 
        & 0.9299& -0.44 & 0.9357&+0.14 & 17 \\
        Magnolia  & 0.9497 
        & 0.9625& +1.28 & 0.9656&+1.59 & 34 \\
        Monimiaceae  & 0.9356 
        & 0.9506& +1.50 & 0.9531&+1.75 & 37 \\
        Rubus  & 0.9185 
        & 0.9485& +3.00 & 0.9504&+3.19 & 22 \\
        Ulmus  & 0.8995 & 0.9360& +3.65 & 0.9386&+3.91 & 53 \\
        \hline
    \end{tabular}
\caption{IoU scores between UNet, PlantSAM1, and PlantSAM2 across the plant taxa. The "Delta" columns indicate the percentage difference in IoU score of PlantSAM1 and PlantSAM2 relative to UNet, highlighting the performance improvement for each taxon. The last column indicates the number of used images for each taxon.}
\label{tab:comparaison}
\end{table}


\begin{table}[H]
\centering
    \begin{tabular}{lcccccc}
        \toprule
        \textbf{Taxon}  & \textbf{UNet} 
        & \textbf{PlantSAM1} 
        & \textbf{Delta} 
        & \textbf{PlantSAM2}&\textbf{Delta}& \textbf{Number of Images}\\
        \midrule
        Amborella  & 0.9619& 0.9662& +0.43 &0.9750& +1.31 & 05 \\
        Castanea  & 0.9772& 0.9706& +0.66 &0.9788& +0.16 & 36 \\
        Convolvulaceae  & 0.9381& 0.9181& -2.00 &0.9447& +0.66 & 25 \\
        Desmodium  & 0.9410& 0.9344& -0.66 &0.9528& +1.18 & 11 \\
        Eugenia  & 0.9668& 0.9566& -1.02 &0.9682& +0.14 & 40 \\
        Laurus  & 0.9795& 0.9752& -0.43 &0.9817& +0.22 & 53 \\
        Litsea  & 0.9766& 0.9603& -1.63 &0.9692& -0.74 & 17 \\
        Magnolia  & 0.9822& 0.9769& -0.53 &0.9836& +0.14 & 34 \\
        Monimiaceae  & 0.9775& 0.9698& -0.77 &0.9775& +0.00 & 37 \\
        Rubus  & 0.9714& 0.9711& -0.03 &0.9798& +0.84 & 22 \\
        Ulmus  & 0.9654& 0.9600& -0.54 &0.9729& +0.75 & 53 \\
        \hline
    \end{tabular} 
\caption{Dice scores between UNet, PlantSAM1, and PlantSAM2 across the plant taxa. The "Delta" columns indicate the percentage change in DICE score of PlantSAM1 and PlantSAM2 relative to UNet, highlighting the performance improvement for each taxon. The last column indicates the number of used images for each taxon.}
\label{tab:comparaison-dice}
\end{table}


\begin{figure}[]
    \centering
    \includegraphics[width=.7\linewidth]{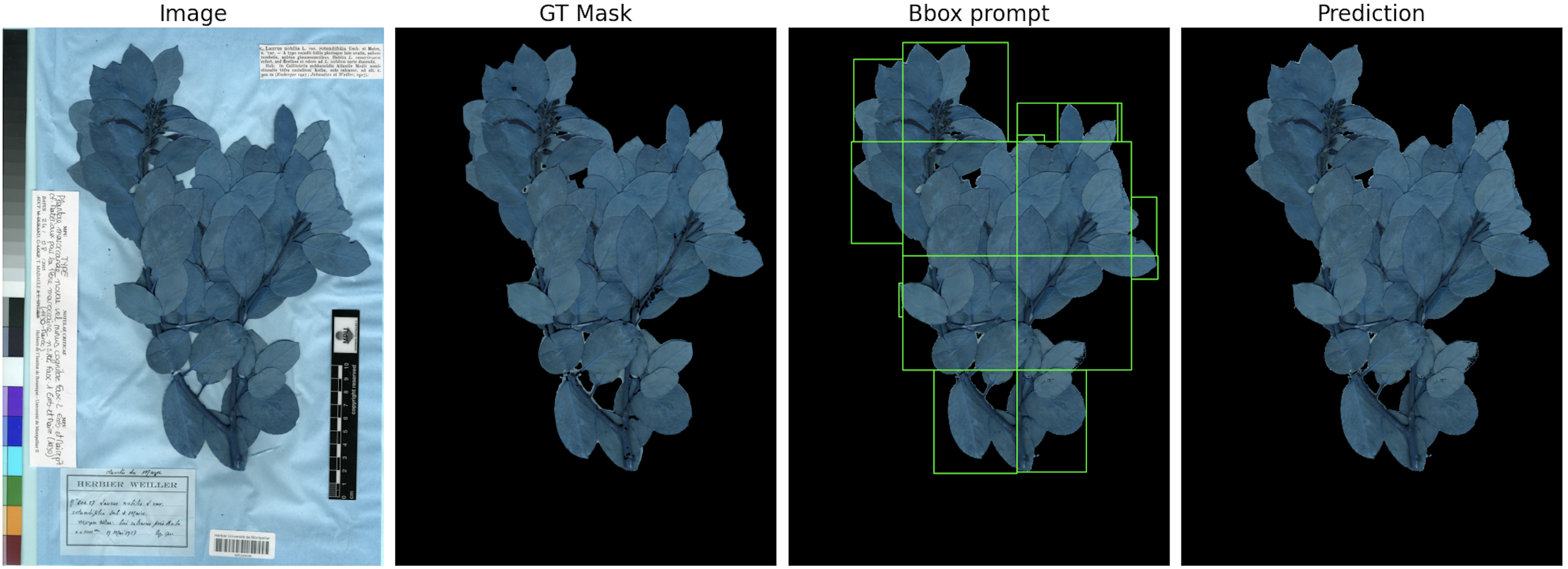}
    \includegraphics[width=.7\linewidth]{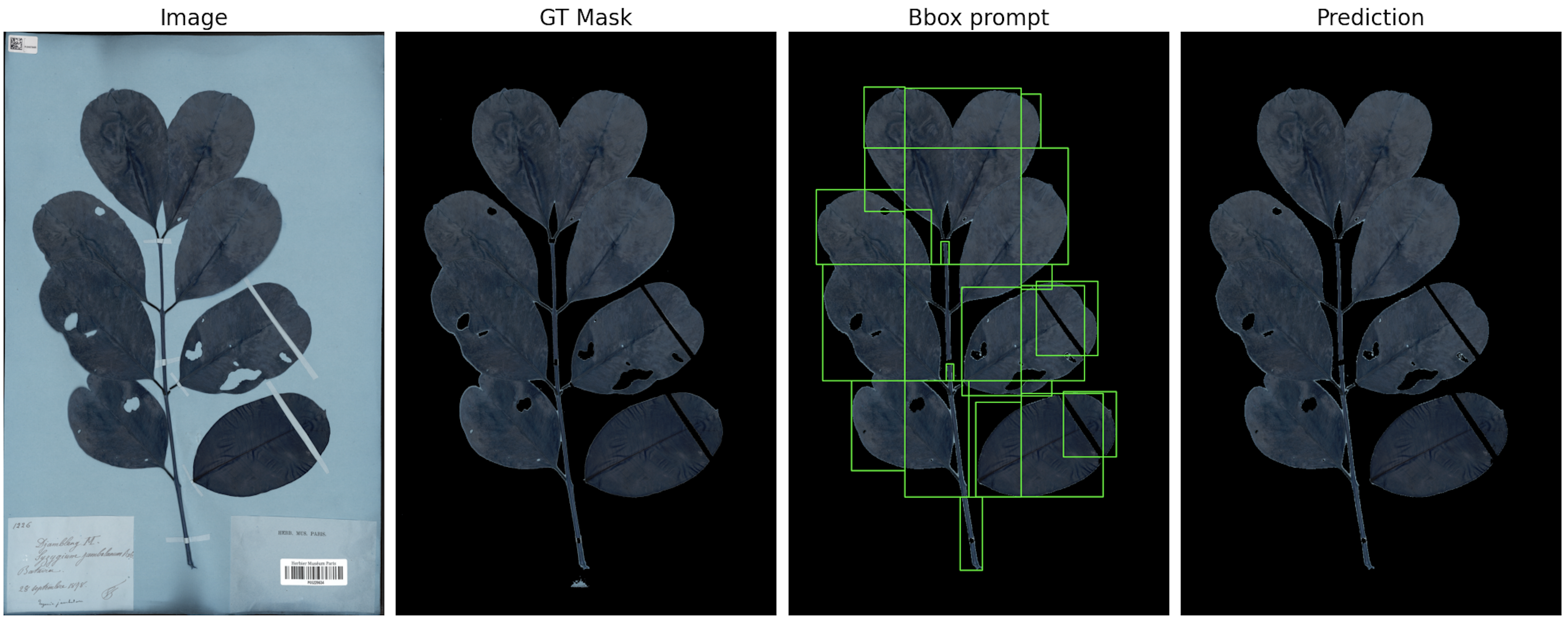}
    \includegraphics[width=.7\linewidth]{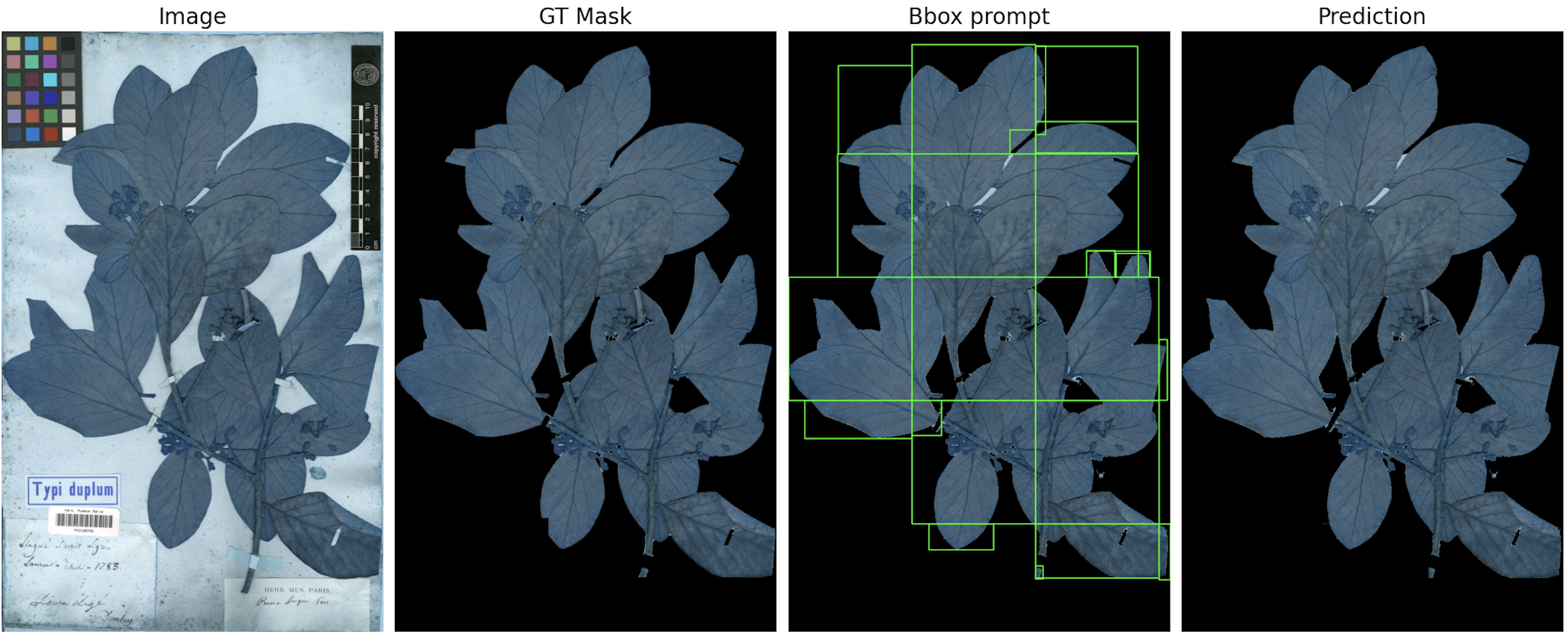}
    \caption{Illustration of the bounding box pipeline, showing the progression from the original image to the ground truth mask (GT Mask), bounding box-based prompts (Bbox prompt), and the final segmentation output (usable mask).}
    \label{fig:combined}
\end{figure}

Figure~\ref{fig:combined} shows a visual illustration of the masks produced by the pipeline. The first column of the figure corresponds to the original image, while the second column displays the manually annotated ground truth mask (GT Mask). The third column presents the initial masks generated by the pipeline. The final column demonstrates the fully segmented masks (a usable mask), where bounding box information has been integrated to produce a complete delineation of the visible plant regions. This pipeline emphasizes the pipeline’s alignment with the ground truth, effectively isolating plant structures from their backgrounds.

\subsubsection{Generalization to Challenging Conditions}

\begin{figure}[]
    \centering
    \begin{subfigure}{0.3\linewidth}
        \centering
        \includegraphics[width=1\linewidth, height=4cm]{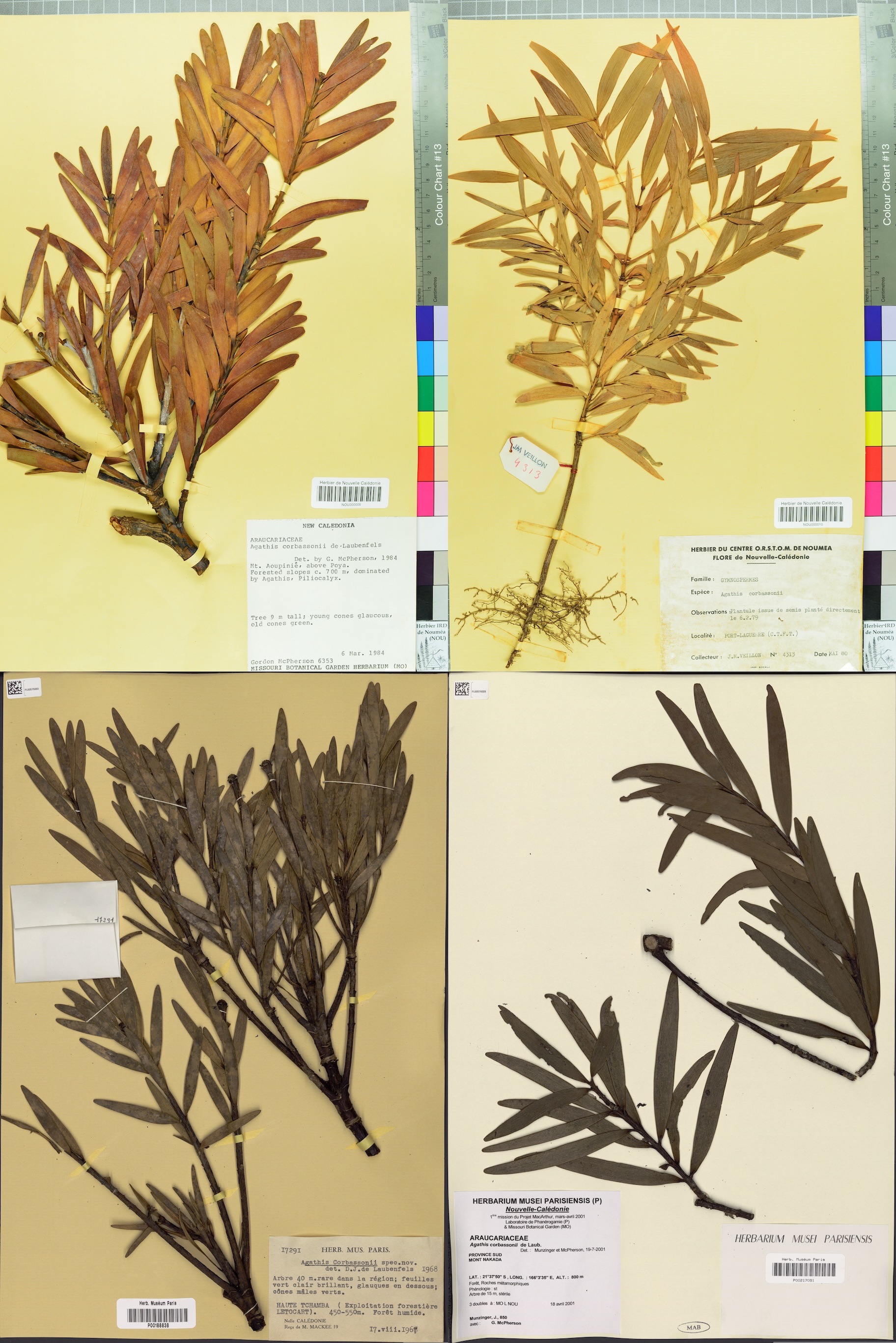}
        \caption{Slender Leaves on Pale Background}
        \label{fig:mosaic2}
    \end{subfigure}
    \hfill
    \begin{subfigure}{0.3\linewidth}
        \centering
        \includegraphics[width=1\linewidth, height=4cm]{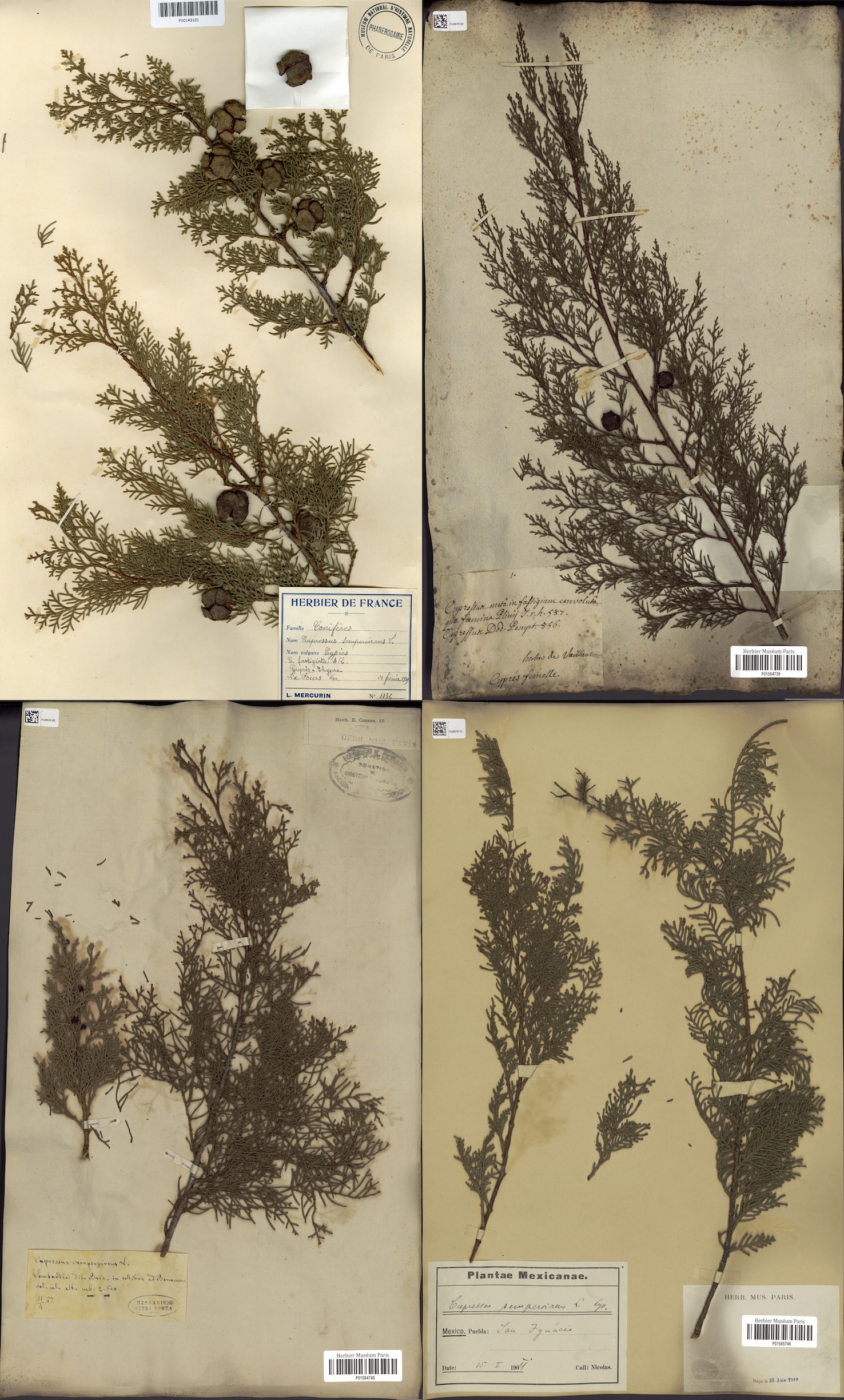}
        \caption{Thin armatures} 
        \label{fig:mosaic3}
    \end{subfigure}
    \hfill
    \begin{subfigure}{0.3\linewidth}
        \centering
        \includegraphics[width=1\linewidth, height=4cm]{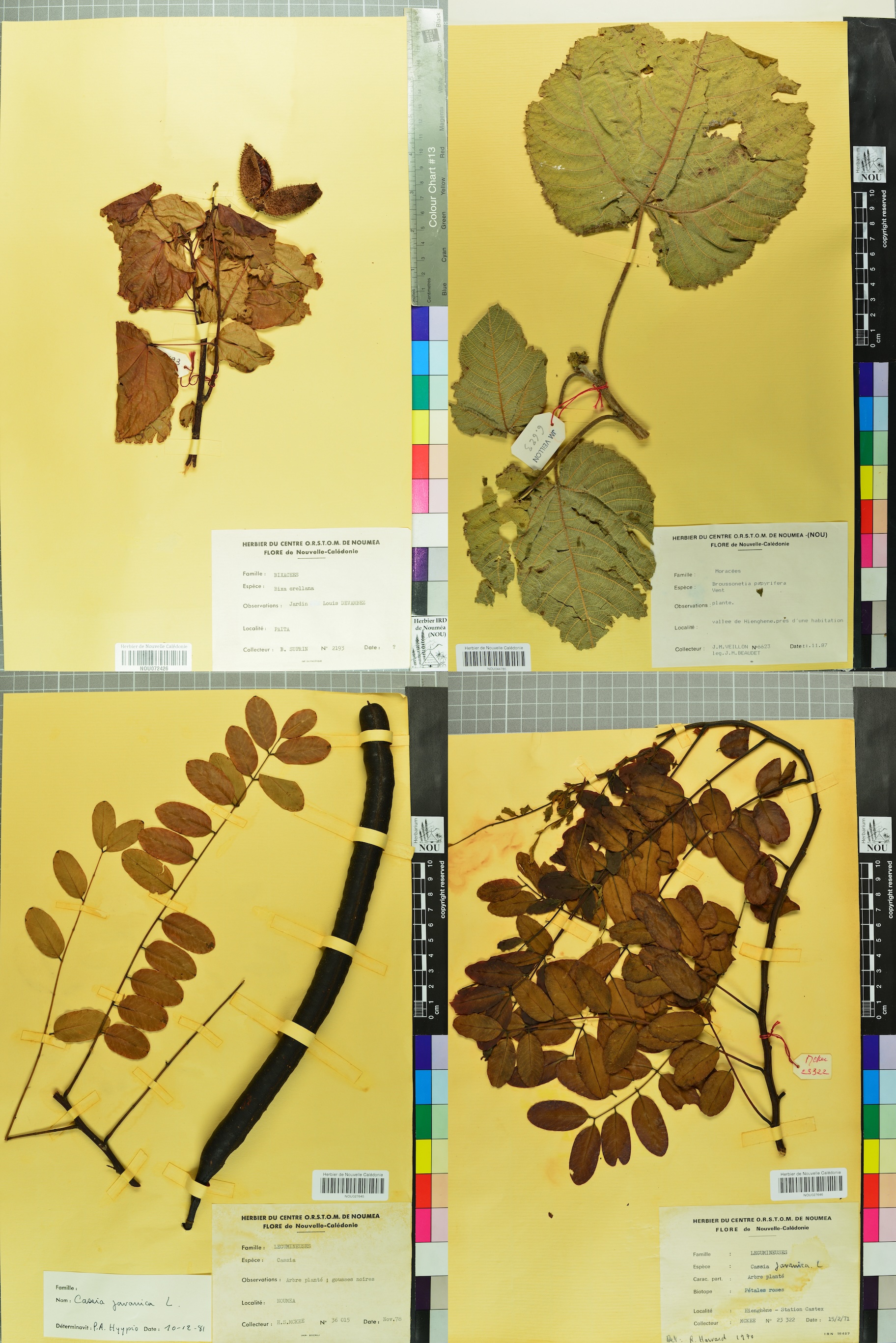}
        \caption{Yellow}
        \label{fig:mosaic6}
    \end{subfigure}
    
    \vspace{4em}
    
    \begin{subfigure}{0.3\linewidth}
        \centering
        \includegraphics[width=.8\linewidth]{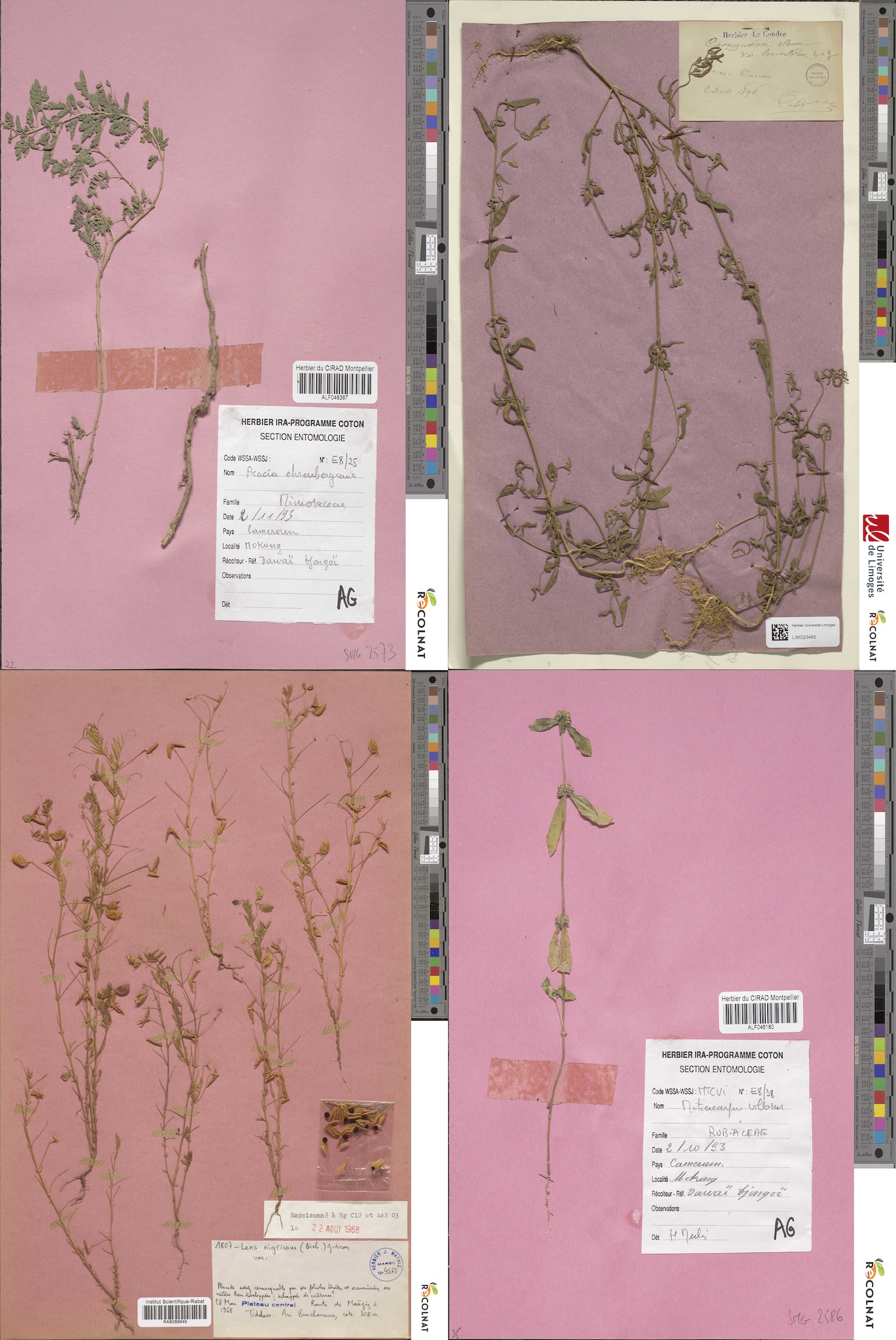}
        \caption{Pink}
        \label{fig:mosaic8}
    \end{subfigure}
    \hfill
    \begin{subfigure}{0.3\linewidth}
        \centering
        \includegraphics[width=.8\linewidth]{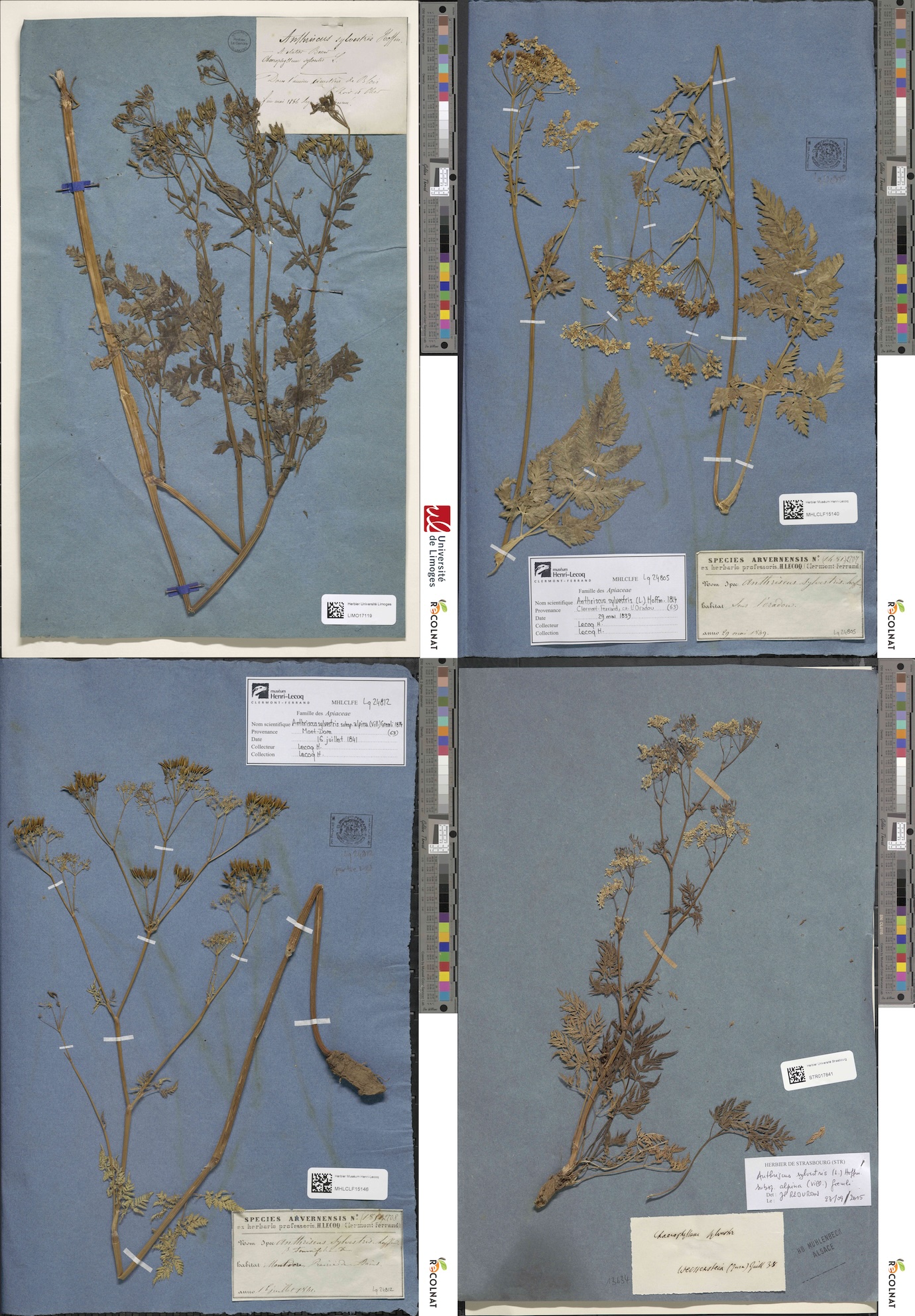}
        \caption{Dark grey}
        \label{fig:mosaic4}
    \end{subfigure}
    \hfill
    \begin{subfigure}{0.3\linewidth}
        \centering
        \includegraphics[width=.8\linewidth]{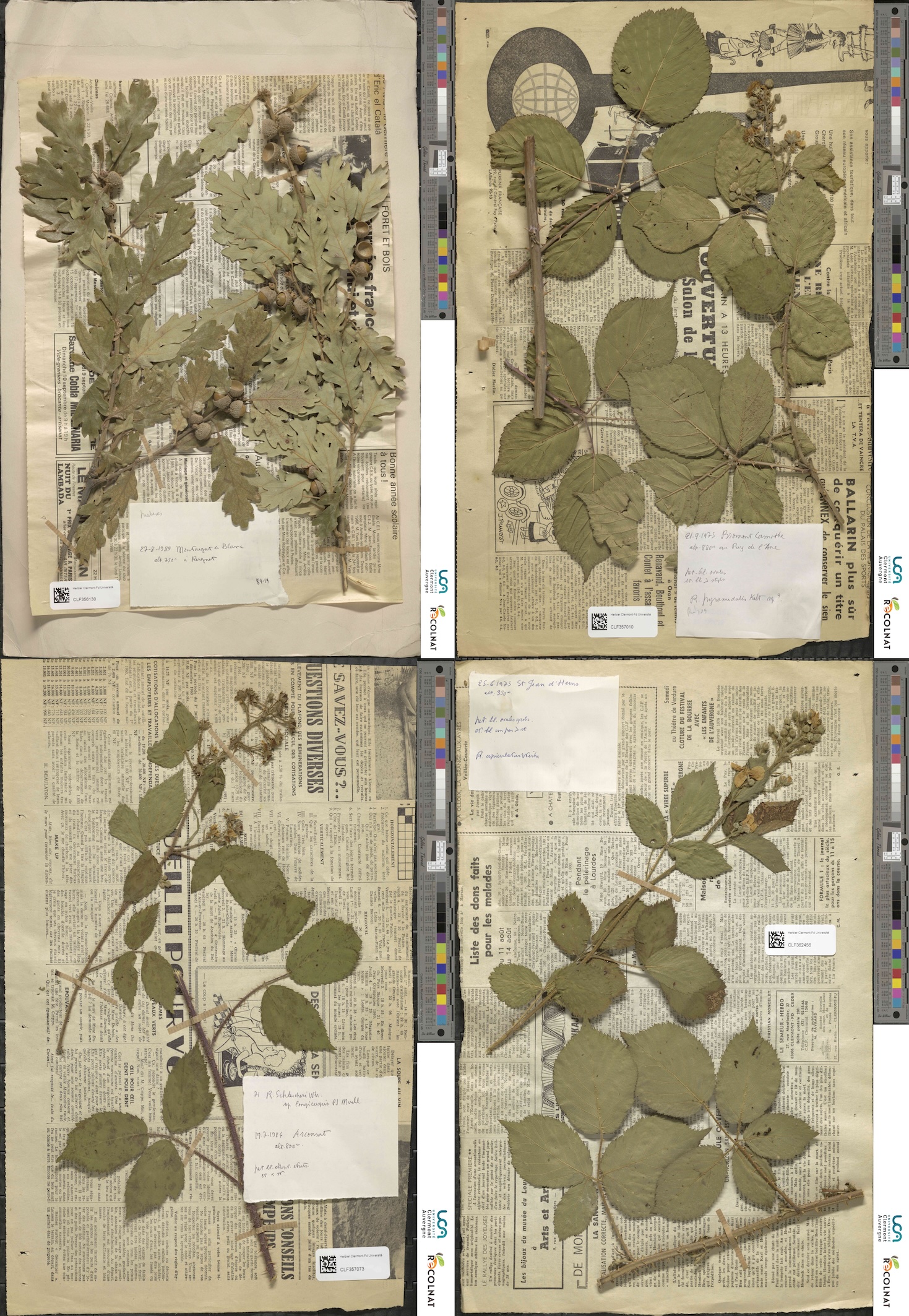}
        \caption{News paper}
        \label{fig:mosaic9}
    \end{subfigure}
    \caption{Examples of herbarium images with various background types, illustrating the diversity in texture and color found across the dataset. The backgrounds include (a) Slender Leaves on Pale Background, (b) Aged paper, (c) Yellow, (d) Pink, (e) Dark grey, and (f) Newspaper. This variability in background characteristics poses challenges for segmentation and highlights the importance of robust preprocessing and model adaptability.}
    \label{fig:mosaics}
\end{figure}

SAM2’s generalization to challenging conditions was evaluated on a distinct Out Of Distribution (OOD) dataset \cite{castanet2025digitised} of 171 images without annotations, featuring noisy or colored backgrounds, intricate plant structures, and artifacts such as pins or overlapping components. This dataset was used to assess the model’s ability to handle challenging conditions. As illustrated in Figure~\ref{fig:mosaics}, the dataset includes a variety of background textures and colors, such as mosaic patterns, yellow, pink, dark grey, and newspaper backgrounds. These variations highlight the segmentation challenges posed by non-uniform backgrounds and emphasize the importance of robust preprocessing and adaptable models for achieving accurate segmentation results.

\begin{figure}[]
    \centering
    \begin{subfigure}[t]{0.495\linewidth}
        \centering
       \includegraphics[width=.5\linewidth]{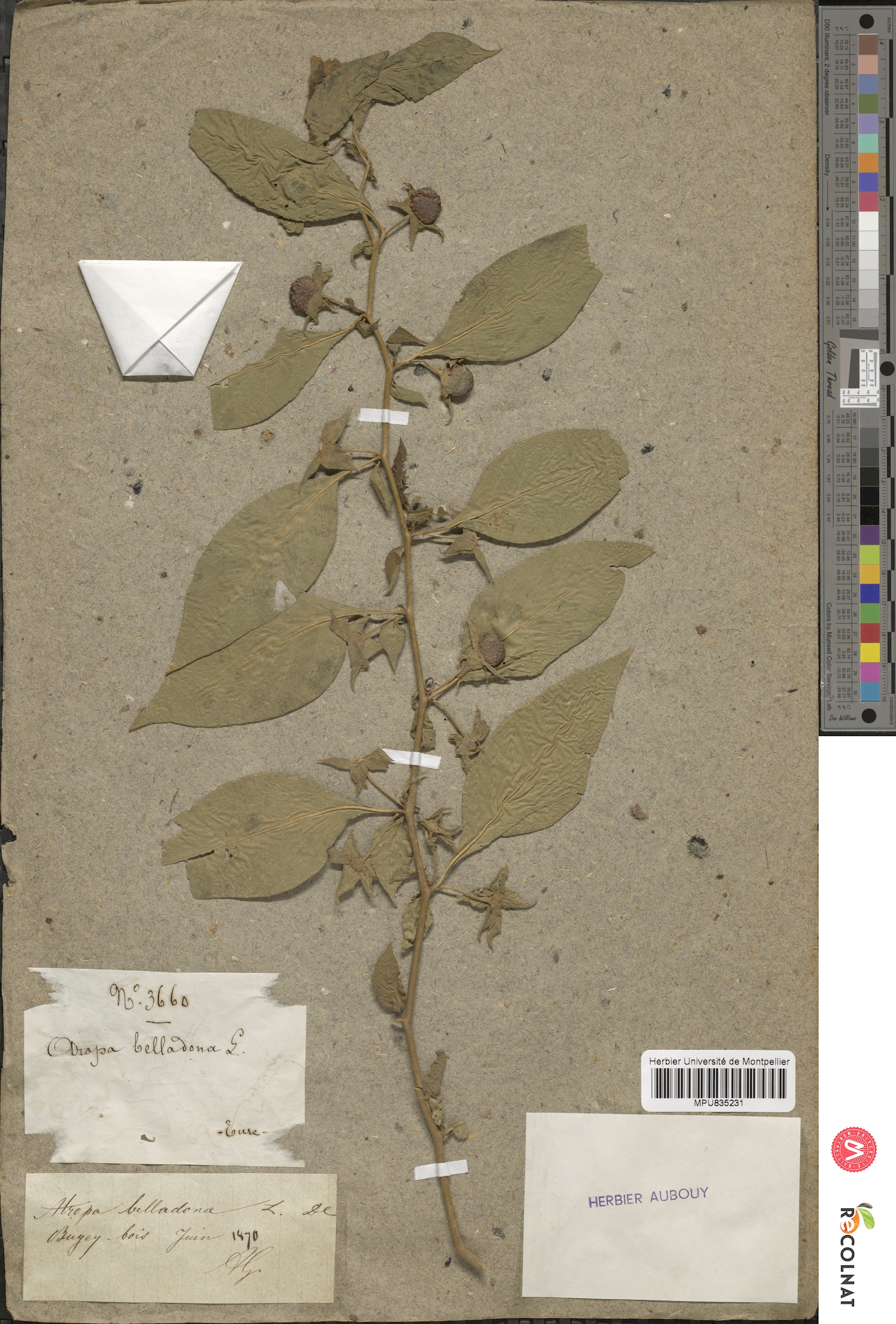}
        \caption{Original image}
    \end{subfigure}
    \begin{subfigure}[t]{0.495\linewidth}
        \centering
        \includegraphics[width=.5\linewidth]{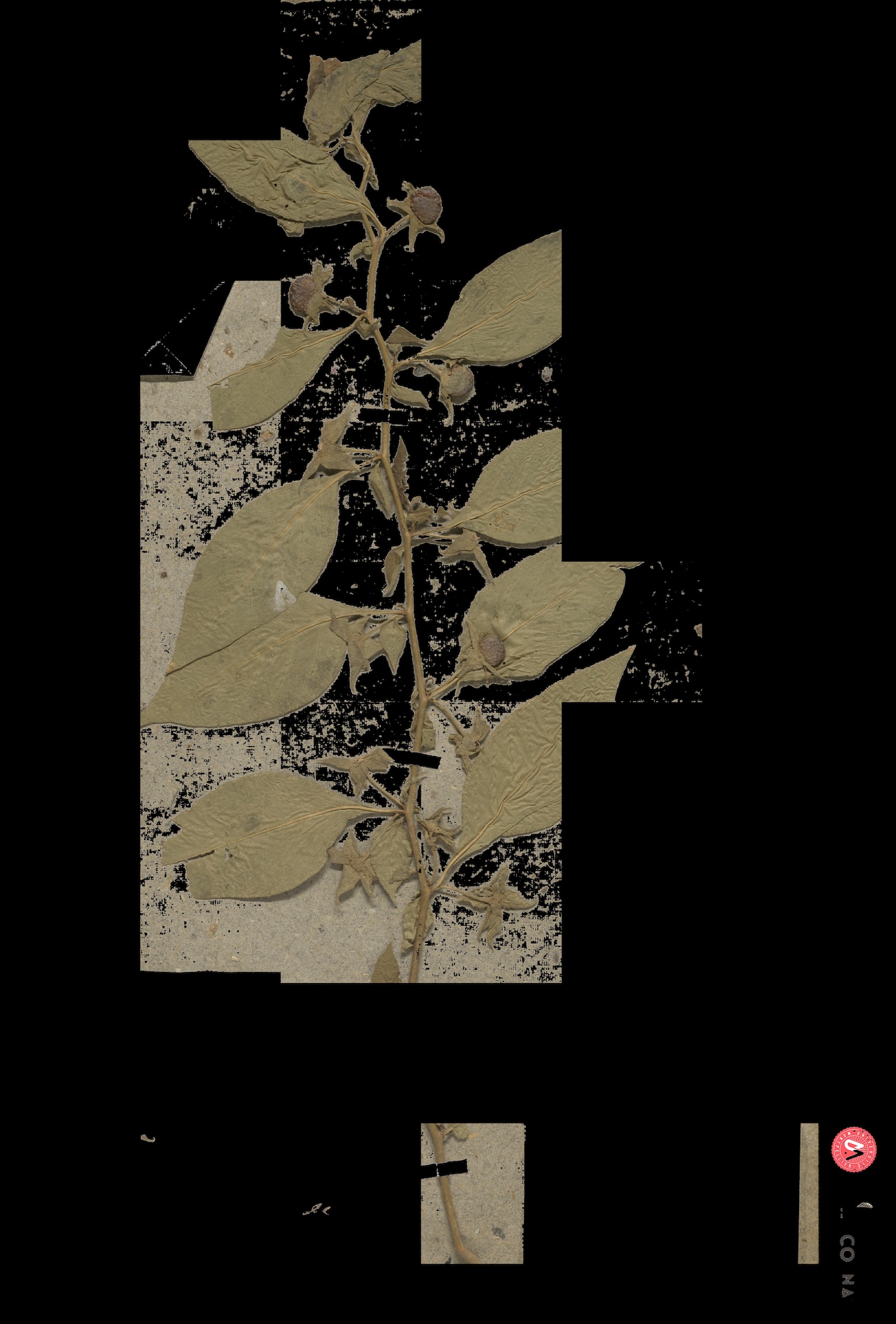}
        \caption{Segmented image}
    \end{subfigure}
    \caption{Example of an unusable mask resulting from challenging conditions. The image on the left shows the original image, while the image on the right shows the segmented version. These challenges include poorly defined contours, low contrast between the plant and background, missing plant elements, and the inclusion of non-plant background components.}
    \label{fig:example-incomplete-segmentation}
\end{figure}

The results, detailed in Tables~\ref{tab:not_usable_masks} and ~\ref{tab:usable_masks}, highlight PlantSAM2’s ability to generate usable masks in over 75\% of cases involving challenging conditions, such as thin armatures (e.g., spines, prickles, or thorns) or colored backgrounds. A usable mask is a segmentation output that, based on visual evaluation, accurately captures the plant structures with minimal background inclusion or artifacts. It correctly delineates the boundaries of the plant regions, preserving fine details necessary for downstream tasks such as species identification or morphological analysis (See Figure \ref{fig:combined} for an example of a usable mask). In contrast, UNet struggled significantly, producing higher proportions of unusable masks. An unusable mask is a segmentation output that, based on visual evaluation, fails to accurately capture the plant structures. This often results from excessive background inclusion, missed plant parts, or segmentation errors caused by artifacts like pins or overlapping components, compromising the quality and reliability of downstream analyses  (See Figure \ref{fig:example-incomplete-segmentation} for an example of a usable mask).


\begin{table}[]
    \centering
    \begin{tabular}{lccc}
        \toprule
        \textbf{Challenging condition} & \multicolumn{2}{c}{\textbf{Percentage of unusable masks (\%)}} & \textbf{Number of Images} \\
        \cmidrule(r){2-3} 
        \textbf{} & \textbf{UNet} & \textbf{PlantSAM2} & \textbf{} \\
        \midrule
        Blue background         & 90.91     & 54.55   & 11 \\
        Grey background         & 67.27     & 29.09   & 55 \\
        Yellow background       & 48.28     & 00.00   & 29 \\
        Brown background        & 100.0     & 53.33   & 15 \\
        Orange background       & 33.33     & 00.00   & 03 \\
        Pink background         & 40.00     & 40.00   & 05 \\
        Thin Armatures          & 07.69     & 03.85   & 26 \\
        Long Thin Armatures     & 41.67     & 08.33   & 12 \\
        Long Leaves             & 50.00     & 00.00   & 04 \\
        Pins                    & 18.18     & 00.00   & 11 \\
        \hline
    \end{tabular}
    \caption{Percentage of unusable masks generated by UNet and PlantSAM2 for various challenging conditions. The last column indicates the number of images analyzed in each category, highlighting PlantSAM2’s improved ability to reduce unusable masks compared to UNet.}
    \label{tab:not_usable_masks}
\end{table}

\begin{table}[]
    \centering
    \begin{tabular}{lccc}
        \toprule
        \textbf{Challenging condition} & \multicolumn{2}{c}{\textbf{Proportion of usable masks (\%)}} & \textbf{Number of Images} \\
        \cmidrule(r){2-3} 
        \textbf{} & \textbf{UNet}           & \textbf{PlantSAM2}      & \textbf{} \\
        \midrule
        Blue background            & 00.00     & 09.09  & 11 \\
        Grey background            & 05.45     & 34.55  & 55 \\
        Yellow background          & 10.34     & 65.52  & 29 \\
        Brown background           & 00.00     & 26.67  & 15 \\
        Orange background          & 33.33     & 33.33  & 03 \\
        Pink background            & 00.00     & 00.00  & 05 \\
        Thin Armatures             & 19.23     & 61.54  & 26 \\
        Long Thin Armatures        & 25.00     & 58.33  & 12 \\
        Long Leaves                & 25.00     & 75.00  & 04 \\
        Pins                       & 09.09     & 90.91  & 11 \\
        \hline
    \end{tabular}
    \caption{Proportion of usable masks generated by UNet and PlantSAM2 across various challenging conditions. The last column indicates the number of images analyzed for each category, highlighting PlantSAM2’s superior ability to produce usable masks in most scenarios.}
    \label{tab:usable_masks}
\end{table}


Table~\ref{tab:performance_comparison} further provides a detailed comparison of the percentage of best-performing masks across various  challenging conditions. PlantSAM2 consistently outperformed UNet across nearly all conditions, with its performance being particularly prominent in complex scenarios such as "Thin Armatures" (73.08\%) and "Long Thin Armatures" (75\%). For challenging background colors like "Yellow" and "Grey" PlantSAM2 achieved best-mask rates of 89.66\% and 69.09\%, respectively. In contrast, UNet’s performance remained limited, rarely surpassing 25\% in most  challenging conditions. Notably, PlantSAM2 exhibited excellent robustness in handling artifacts such as pins, achieving a success rate of 90.91\%. Additionally, the  column "NONE" reflects instances where neither model succeeded, highlighting the remaining challenges in dealing with particularly difficult cases, such as brown and grey backgrounds. The last column provides context on the number of images analyzed for each condition, further demonstrating the robustness and scalability of PlantSAM2 across diverse datasets.

\begin{table}[]
    \centering
    \begin{tabular}{llccc}
        \toprule
        \textbf{Challenging condition} & \multicolumn{3}{c}{\textbf{Percentage of best-performing masks (\%)}} & \textbf{Number of Images} \\
        \cmidrule(r){2-4} 
        \textbf{} & \textbf{UNet} & \textbf{PlantSAM2} & \textbf{None} & \textbf{} \\
        \midrule
        Blue background           & 09.09      & 72.73   & 18.18   & 11 \\
        Grey background           & 07.27      & 69.09   & 23.64   & 55 \\
        Yellow background         & 10.34      & 89.66   & 00.00   & 29 \\
        Brown background          & 20.00      & 46.67   & 33.33   & 15 \\
        Orange background         & 33.33      & 66.67   & 00.00   & 03 \\
        Pink background           & 60.00      & 40.00   & 00.00   & 05 \\
        Thin Armatures            & 23.08      & 73.08   & 03.85   & 26 \\
        Long Thin Armatures       & 25.00      & 75.00   & 00.00   & 12 \\
        Long Leaves               & 25.00      & 75.00   & 00.00   & 04 \\
        Pins                      & 09.09      & 90.91   & 00.00   & 11 \\
        \hline
    \end{tabular}
    \caption{Comparison of model performance across various challenging conditions, highlighting the percentage of best-performing masks produced by UNet, PlantSAM2, or no model (NONE). The last column indicates the number of images analyzed for each condition.}
    \label{tab:performance_comparison}
\end{table}

Figure~\ref{fig:example-incomplete-segmentation} illustrates a typical failure case, where low contrast and poorly defined plant contours hindered segmentation. Despite these challenges, PlantSAM2 reduced unusable masks by over 50\% compared to UNet, demonstrating its robustness across diverse scenarios.

It is important to interpret these results cautiously, as the datasets used to compute IoU and Dice scores are not perfect representations of ideal segmentation masks. While they are close to the ground truth, we sometimes observe segmentation traits produced by PlantSAM2 that could be considered improvements over the provided annotations. However, these enhancements could result in lower IoU scores instead of higher ones, highlighting a possible limitation of current evaluation metrics. To address these discrepancies and better understand PlantSAM's segmentation performance, we propose further analysis using classification  techniques.

\subsection{Impact of segmentation on classification}

To further evaluate the impact of segmentation on deep learning models for herbarium specimens, we assessed its effect on classification performance. A key challenge in training classification models on herbarium images is the high proportion of background relative to plant structures (Figure \ref{fig:heat-maps-plant-species}). As shown in Table~\ref{tab:species_coverage}, non-plant elements can account for over 90\% of an image, leading models to learn misleading features. By applying segmentation with PlantSAM, we isolate relevant botanical structures, reducing background influence. Table~\ref{tab:dataset_distribution} presents the distribution of images in the training and validation datasets for five botanical traits: armatures, fruits, acuminate leaf tips, infructescence, and acute leaf bases. The dataset consists of manually annotated herbarium images, where each specimen is labeled for one or more traits.


\begin{table*}
\centering
\caption{Distribution of images in the training and validation datasets for the 5 studied traits for the classification downstream task.}\label{tab:datasets-distribution}
\begin{tabular}{lllll}
\toprule
Trait                       & Train & Validation  \\
\midrule
Armatures                      & 1607  & 344  \\
Fruits                      & 1623  & 335  \\
Leaves with acuminate tips  & 1479  & 305  \\
Infructescence              & 1394  & 302  \\
Leaves with an acute base   & 1219  & 256  \\
\hline
\end{tabular}
\label{tab:dataset_distribution}
\end{table*}

\begin{figure}[H]
    \centering
     \fbox{\includegraphics[width=.8\linewidth]{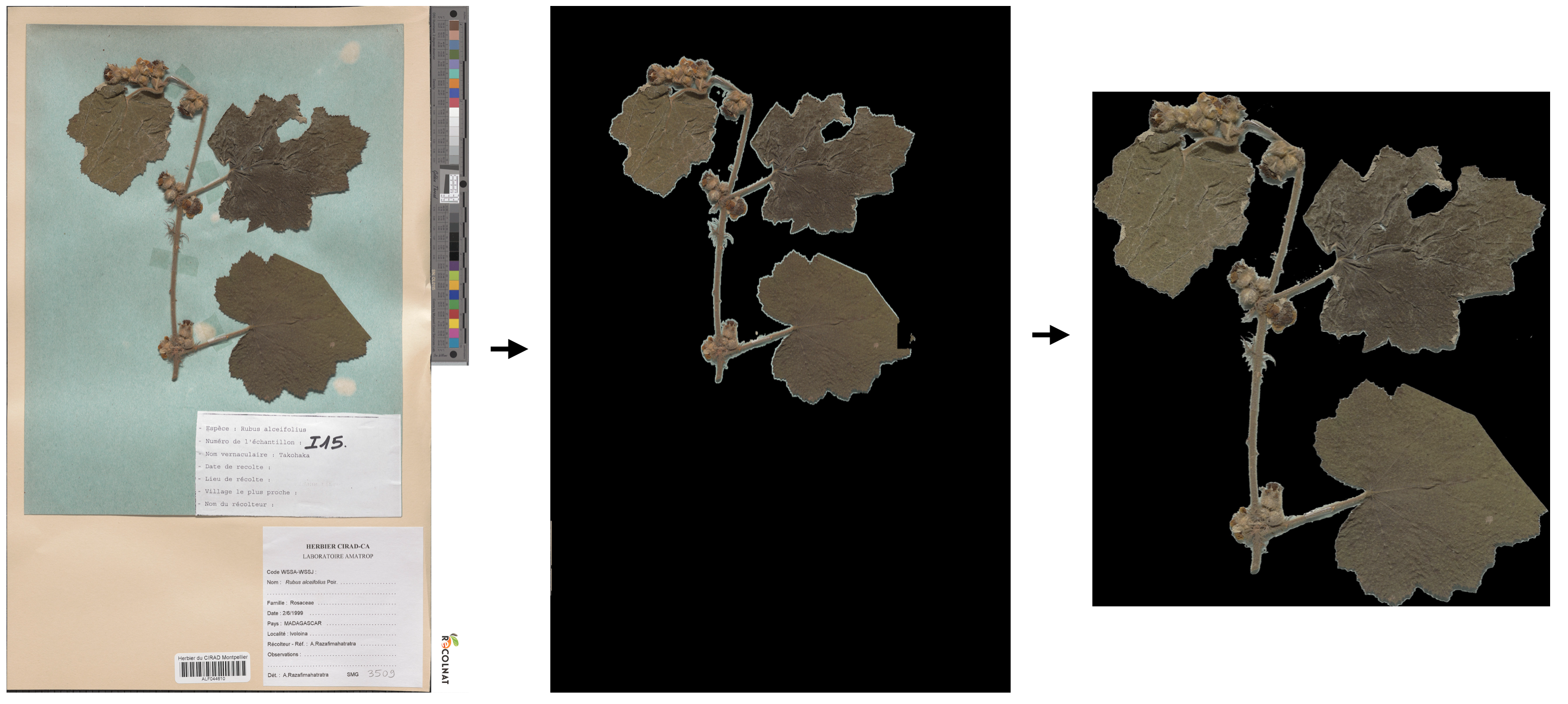}}
    \caption{Illustration of the segmentation-based cropping process. \textbf{Left:} Original herbarium sheet with background elements. \textbf{Middle:} Segmented image isolating the plant structures. \textbf{Right:} Cropped segmented image, removing unnecessary empty space and improving resolution for classification.}
    \label{fig:cropping_illustration}
\end{figure}

We trained ResNet101 model on three versions of the dataset: i) unsegmented images (raw images with background); ii) segmented images (background removed using PlantSAM); iii) segmented cropped images (only plant regions retained). A major challenge in standard deep learning pipelines is that herbarium images are typically resized to a fixed dimension (e.g., 224×224). This resizing operation compresses plant features along with large empty background regions, leading to a loss of fine morphological details. By using segmentation-based cropping, we eliminate unnecessary empty space, allowing for higher-resolution plant structures to be preserved within the same computational constraints. Figure~\ref{fig:cropping_illustration} illustrates this process. The original herbarium sheet (left) contains large non-plant regions, including labels and background textures. The segmentation step (middle) isolates the plant, and the cropping step (right) removes empty space, ensuring that only the essential botanical components remain.


Table~\ref{tab:classification_performance} summarizes the classification performance. Accuracy (Acc) and F1-score (F1) are reported, with delta values indicating the improvements relative to unsegmented images. The results show that segmentation improves classification accuracy across all traits. Cropped segmented images outperform both raw and segmented images, especially for traits with fine morphological structures such as armatures (+4.36\%). F1-scores also show consistent improvements, indicating that segmentation not only enhances accuracy but also increases precision and recall. The highest improvement is observed for armatures, where segmentation eliminates background noise that could mislead the model. These results demonstrate that removing background elements leads to more discriminative features, making it easier for deep learning models to classify plant traits.


\begin{table*}
\centering
\caption{Performance evaluation of the ResNet101 model on unsegmented and segmented images across five botanical traits. The metrics used for comparison are accuracy (Acc) and F1-Score (F1). The delta values represent the performance difference between segmented and unsegmented images.}\label{tab:resnet_complete_performances}
\begin{tabular}{lllllllllll} 
\toprule
Trait                       & \multicolumn{2}{c}{unsegmented} & \multicolumn{2}{c}{segmented} & \multicolumn{2}{c}{delta} & \multicolumn{2}{c}{segmented cropped} & \multicolumn{2}{c}{delta}  \\ 
\cmidrule(r){2-11} 
                            & Acc   & F1        & Acc   & F1        & Acc    & F1      & Acc   & F1       & Acc    & F1  \\  \midrule
Armatures                      & 90.41 & 89.18     & 92.73  & 89.77    & +2.32  & +0.59   & 94.77 & 91.56    & +4.36  & +2.38  \\
Fruits                      & 57.01 & 62.02     & 58.21  & 64.32    & +1.2	 & +2.3    & 60.00 & 64.41    & +2.99  & +2.39  \\
Leaves with acuminate tips  & 69.84 & 70.24     & 70.16  & 71.47    & +0.32  & +1.23   & 72.46 & 73.43    & +2.62  & +3.19  \\
Infructescence              & 58.94 & 59.56     & 60.26  & 62.98    & +1.32  & +3.42   & 59.93 & 63.71    & +0.99  & +4.15  \\
Leaves with an acute base   & 68.75 & 68.24     & 69.18  & 69.57    & +0.43  & +1.33   & 69.53 & 69.75    & +0.78  & +1.51   \\
\hline
\end{tabular}
\label{tab:classification_performance}
\end{table*}

\subsection{Semi-Automatic Annotation Tool for Refining Segmentation Masks}

\begin{figure}[H]
    \centering
    \fbox{\includegraphics[width=.7\linewidth]{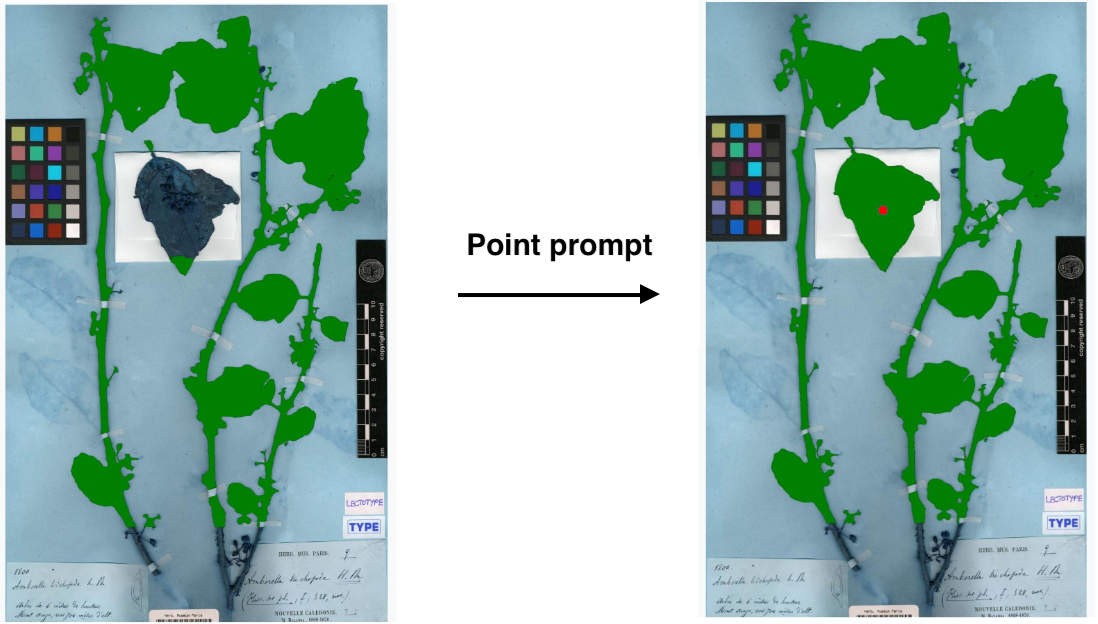}}
    \caption{Example of improving segmentation using a point prompt in the developed application. The user indicates the region to be segmented by providing a point prompt (shown as a red dot on the image to the right). This interaction refines the recoverable mask by correcting segmentation errors and focusing on the specific regions of interest, resulting in a more accurate mask.}
    \label{fig:enter-label}
\end{figure}

Our pipeline demonstrates strong potential for integration into a semi-automatic annotation tool. While fully automatic segmentation remains challenging for some cases, this limitation can be addressed through a semi-automatic approach where users refine the masks by providing precise point prompts to SAM. We have developed an application that incorporates PlantSAM2 pipeline, enabling the semi-automatic correction of masks initially deemed unusable. This process relies on direct interaction with expert users who guide SAM2 to produce more accurate segmentation results (Figure \ref{fig:enter-label}).


The application serves two main purposes: correcting poor segmentations produced by the pipeline PlantSAM2 and enhancing the segmentation of complex cases to expand the training dataset for fine-tuning the SAM2 model. Users can upload their images, which are initially processed by PlantSAM2. The images are then presented one by one, allowing users to interact with specific regions by providing point prompts to SAM2. This interaction transforms unusable masks into usable ones.

\section{Discussion \& Conclusions}

In this work, we developed an automatic segmentation pipeline tailored for herbarium specimens, leveraging object detection techniques to integrate the strengths of the Segment Anything Model (SAM) while addressing its limitations. By combining SAM with YOLOv10, we created a streamlined pipeline that significantly outperformed traditional models like UNet. The results demonstrated PlantSAM2’s consistent superiority over both UNet and PlantSAM1 in terms of segmentation accuracy (IoU and Dice scores) and its ability to generalize to challenging scenarios, such as colored backgrounds and intricate plant regions.

Beyond segmentation, we demonstrated the substantial impact of background removal on classification performance. By isolating plant structures and eliminating irrelevant background elements, segmented images improved the accuracy and F1-score of classification models across all studied botanical traits. Additionally, the segmentation-based cropping strategy allowed for better utilization of image resolution by focusing on relevant plant regions, preventing the loss of fine morphological details during preprocessing. These improvements reinforce the importance of segmentation as a preprocessing step for deep learning applications in herbarium images analysis.

This pipeline not only represented a technical improvement but also provided practical advancements in herbarium image processing. The integration of a semi-automatic annotation application empowered users to efficiently refine masks, converting otherwise unusable masks into usable ones with minimal manual intervention. This approach could accelerate the image annotation process, enabling the rapid generation of new datasets while substantially reducing the time and labor typically required for manual annotation.

In conclusion, the combination of SAM2 with an object detection model and semi-automatic refinement tools has expanded the scope of treatable herbarium images. Future extensions of this pipeline could include adaptive prompts to further enhance segmentation in challenging cases or multi-modal techniques that integrate textual and morphological data. These advancements would support the development of increasingly sophisticated tools for herbarium data analysis, fostering broader applications in taxonomy, conservation biology, and biodiversity research.

\section*{Acknowledgements}
This work was partly funded by the French National Research Agency in the context of the e-Col+ project (ANR-21-ESRE-0053). This project was provided with computing HPC and storage resources through the 2023-A0150114385 grant by GENCI at IDRIS.

\bibliographystyle{plain} 
\bibliography{references}

\end{document}